\documentclass[10pt]{article}
\usepackage[table]{xcolor}
\usepackage[preprint]{tmlr}

\usepackage{amsmath,amsfonts,bm}

\def\eqref#1{equation~\ref{#1}}

\def\1{\bm{1}}

\def\vtheta{{\bm{\theta}}}
\def\vomega{{\bm{\omega}}}
\def\vphi{{\bm{\phi}}}

\def\vx{{\bm{x}}}
\def\vy{{\bm{y}}}

\def\mX{{\bm{X}}}

\DeclareMathAlphabet{\mathsfit}{\encodingdefault}{\sfdefault}{m}{sl}
\SetMathAlphabet{\mathsfit}{bold}{\encodingdefault}{\sfdefault}{bx}{n}

\def\gB{{\mathcal{B}}}
\def\gC{{\mathcal{C}}}

\def\gE{{\mathcal{E}}}

\def\gL{{\mathcal{L}}}

\def\gO{{\mathcal{O}}}

\def\gS{{\mathcal{S}}}

\def\gU{{\mathcal{U}}}

\def\gX{{\mathcal{X}}}
\def\gY{{\mathcal{Y}}}

\newcommand{\E}{\mathbb{E}}

\newcommand{\R}{\mathbb{R}}

\newcommand{\softmax}{\mathrm{softmax}}

\DeclareMathOperator*{\argmax}{arg\,max}
\DeclareMathOperator*{\argmin}{arg\,min}

\usepackage{hyperref}
\usepackage{url}
\usepackage[most]{tcolorbox}
\usepackage{nicefrac}
\usepackage{algorithm,algpseudocode}
\usepackage{pifont}
\newcommand{\cmark}{\ding{51}}%
\newcommand{\xmark}{\ding{55}}%
\usepackage{graphicx}
\usepackage{booktabs}
\usepackage{subcaption}
\usepackage{blindtext}
\usepackage{wrapfig}
\usepackage{enumitem}
\usepackage[capitalize]{cleveref}
\usepackage{multirow}
\usepackage{makecell}
\usepackage{mathtools}
\usepackage{dsfont}

\def\methodname{{BoSS}\xspace}

\usepackage{soul}

\definecolor{myboxcolor}{rgb}{0.402,0.402,0.402}
\newtcolorbox{mybox}[1][]{
        enhanced,
        title=#1,
        colback=myboxcolor!3,
        colbacktitle=myboxcolor!3,
        coltitle=black,
        left=4pt,
        right=4pt,
        top=4.5pt,
        bottom=0pt,
        attach boxed title to top left={xshift=8pt, yshift=-7pt},
        boxed title style={frame hidden, size=small, colback=myboxcolor!3},
        sharp corners,
        rounded corners,
        arc=7pt,
}

\title{\methodname: A Best-of-Strategies Selector as an Oracle for Deep Active Learning}

\author{\name Denis Huseljic \email denis.huseljic@uni-kassel.de \\
      \addr Intelligent Embedded Systems \\
      University of Kassel \\
      Kassel, Hesse, Germany
      \AND
      \name Paul Hahn \email paul.hahn@uni-kassel.de \\
      \addr Intelligent Embedded Systems \\
      University of Kassel \\
      Kassel, Hesse, Germany
      \AND
      \name Marek Herde \email marek.herde@uni-kassel.de \\
      \addr Intelligent Embedded Systems \\
      University of Kassel \\
      Kassel, Hesse, Germany
      \AND
      \name Christoph Sandrock \email christoph.sandrock@tuwien.ac.at \\
      \addr Machine Learning Research Unit \\
      TU Wien \\
      Vienna, Vienna, Austria
      \AND
      \name Bernhard Sick \email bernhard.sick@uni-kassel.de \\
      \addr Intelligent Embedded Systems \\
      University of Kassel \\
      Kassel, Hesse, Germany
      }

\begin{document}

\maketitle
\begin{abstract}
Active learning (AL) aims to reduce annotation costs while maximizing model performance by iteratively selecting valuable instances.  While foundation models have made it easier to identify these instances, existing selection strategies still lack robustness across different models, annotation budgets, and datasets. To highlight the potential weaknesses of existing AL strategies and provide a reference point for research, we explore oracle strategies, i.e., strategies that approximate the optimal selection by accessing ground-truth information unavailable in practical AL scenarios. Current oracle strategies, however, fail to scale effectively to large datasets and complex deep neural networks. To tackle these limitations, we introduce the Best-of-Strategy Selector (\methodname), a scalable oracle strategy designed for large-scale AL scenarios.  \methodname constructs a set of candidate batches through an ensemble of selection strategies and then selects the batch yielding the highest performance gain. As an ensemble of selection strategies, \methodname can be easily extended with new state-of-the-art strategies as they emerge, ensuring it remains a reliable oracle strategy in the future. Our evaluation demonstrates that i) \methodname outperforms existing oracle strategies, ii) state-of-the-art AL strategies still fall noticeably short of oracle performance, especially in large-scale datasets with many classes, and iii) one possible solution to counteract the inconsistent performance of AL strategies might be to employ an ensemble‑based approach for the selection.
\end{abstract}

\section{Introduction}\label{sec:intro}
Despite the era of foundation models~\citep{gupte2024revisiting}, most machine learning applications still require carefully annotated domain- or task-specific data~\citep{rauch2024towards}. Active learning (AL)~\citep{settles2009active} aims to reduce annotation costs by prioritizing instances that maximize model performance. Selecting which subset of instances to annotate is determined by a selection strategy, which is the most critical element of AL. However, recent studies show that there is no single selection strategy that outperforms every other alternative across different domains, model architectures, and budgets~\citep{munjal2022towards,luth2024navigating,werner2024a}. AL aims to maximize model performance, yet most strategies do not directly select instances that optimize this goal. Instead, they rely on performance-related heuristics, which can be suboptimal in some scenarios and highly effective in others. Moreover, once chosen, a selection strategy typically remains fixed throughout the entire AL process, limiting the ability to adapt to distribution shifts caused by iteratively annotating new instances. For example, some strategies, such as TypiClust, work well in early stages of AL, but tend to fail in later stages~\citep{hacohen2022active}. This inconsistency underscores the challenge of identifying the instances that yield the greatest performance gains for a given budget. 

Nevertheless, it is possible to conceptualize a selection strategy that approximates an optimal selection using a so-called \emph{oracle strategy} that leverages ground truth information, including instance labels or access to the test data. Although this information is unattainable in real AL applications, such an oracle strategy  provides a useful diagnostic reference for assessing state-of-the-art selection strategies. Comparing the performance gap between selection strategies and the oracle strategy reveals how far these approaches deviate from the ideal one and whether that deviation is concentrated in early, later, or across all cycles. Moreover, analyzing how the oracle selects data may offer new insights for refining existing strategies or guide the development of new, even more effective ones.

However, approximating the optimal selection strategy is inherently challenging due to the combinatorial explosion in finding the best subset, along with the necessity of model retraining. Although some studies have tried to approximate the optimal strategy~\citep{sandrock2023exploring,zhou2021towards,werner2024a}, these methods are feasible only for small-scale models and datasets, mainly due to their high computational demands, which arise from assessing the influence of each instance independently instead of in batches.  While~\citet{sandrock2023exploring} focuses on kernel-based models with tabular data, \citet{zhou2021towards} use only small convolutional and recurrent architectures with small-scale datasets such as Fashion-MNIST~\citep{xiao2017fashion}.  Although a more practical strategy was introduced by \citet{werner2024a} recently, it remains computationally expensive for larger budgets. For example, due to high computational costs, the authors extrapolated results for batch sizes above 500. Overall, current oracle strategies do not scale to more challenging, larger datasets, making it impossible to compare them to state-of-the-art AL selection strategies in these settings.

\begin{wrapfigure}[14]{r}{0.45\linewidth}
    \vspace{-1em}
    \includegraphics[width=\linewidth]{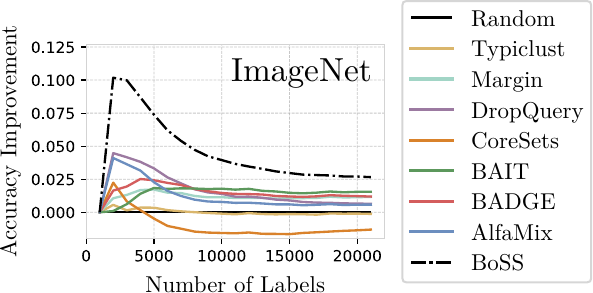}
    \caption{Accuracy improvement over random sampling for \methodname and state-of-the-art selection strategies using DINOv2-ViT-S/14.}\label{fig:ga}
\end{wrapfigure}
In this article, we propose the \emph{Best-of-Strategies Selector (\methodname)}, a simple and scalable oracle strategy for batch AL that approximates the optimal selection and can be efficiently applied to large-scale deep neural networks (DNNs) and datasets. \methodname first constructs a diverse pool of candidate batches through an ensemble of selection strategies. It then adopts a performance-based perspective, selecting the candidate batch that, once annotated, leads to the highest performance improvement. For efficiency, \methodname freezes the pretrained backbone and assesses the performance improvement of candidate batches by retraining only the final layer during selection. By combining a ensemble-based preselection of candidate batches, a performance-based batch assessment, and a frozen backbone, \methodname serves as a batch oracle strategy that also works in large-scale deep AL settings, something that previous oracles do not achieve (cf.~\cref{fig:ga}). Our experiments on a variety of image datasets demonstrate that i) \methodname outperforms existing oracle strategies under comparable computational constraints, ii) current state-of-the-art AL strategies still fall noticeably short of oracle performance, especially in large-scale datasets with many classes, indicating there remains potential for developing stronger strategies, and iii) there is no single AL strategy consistently dominates across all AL cycles, highlighting the potential for a robust ensemble-driven AL strategy.
Our contributions can be summarized as follows:

\begin{mybox}[\textbf{Contributions}]
    \begin{itemize}[leftmargin=*]
    \setlength\itemsep{0pt}
      \item \textbf{Scalable Oracle:} We introduce \methodname, the first batch oracle strategy scalable to large datasets and complex DNNs. \methodname combines an \emph{ensemble of selection strategies} with a \emph{performance-based selection}, efficiently realized by \emph{retraining only the final linear layer}.
      \item \textbf{Comprehensive Evaluation:} Extensive experiments demonstrate that i) \methodname outperforms existing oracle strategies and ii) current state-of-the-art AL strategies fall noticeably short of oracle performance. Our implementation is publicly available at {\url{https://github.com/dhuseljic/dal-toolbox}}.
      \item \textbf{Insights into AL Development:} Our analysis highlights i) that observed gaps between selection strategies and oracle performance suggest room for improvement and ii) that ensemble-based AL approaches can be a potential solution for mitigating the inconsistencies of commonly employed single AL strategies.
      \end{itemize}
\end{mybox}

\section{Related Work}\label{sec:rel_work}
\textbf{Selection strategies} in AL are typically divided into uncertainty- and representativeness-based strategies. While popular uncertainty-based strategies, such as Margin~\citep{settles2009active} or BADGE~\citep{ash2020deep}, assume that a selection of difficult (or uncertain) instances improves performance, representativeness-based strategies, such as Typiclust~\citep{hacohen2022active}, favor instances that best represent the underlying data distribution.  In recent years, a combination of both has proven to work well~\citep{ash2021gone,gupte2024revisiting} because both \emph{heuristics} are partially related to model performance. Furthermore, as it is common to select batches in deep learning, most strategies ensure diversity within a batch through clustering, avoiding the selection of similar instances~\citep{kirsch2023stochastic,ash2020deep,gupte2024revisiting}.

Despite substantial progress in AL, selection strategies can still fail because heuristics such as uncertainty or representativeness do not guarantee performance improvements~\citep{zhao2021uncertaintyaware}. Consequently, recent studies have increasingly focused on evaluating the \textbf{robustness of AL strategies}, revealing significant challenges in identifying the universally best strategy. \citet{munjal2022towards} emphasize this difficulty by demonstrating that no single selection strategy consistently outperforms others, with results heavily dependent on experimental conditions and hyperparameter tuning during AL cycles. Similarly, \citet{luth2024navigating} and \citet{rauch2023activeglae} further explore these inconsistencies and propose a standardized evaluation protocol, finding that BADGE generally performs best across diverse experimental setups. However, in contrast, \citet{werner2024a} evaluate strategies across multiple domains, including images, text, and tabular data, concluding that Margin yields the most consistent performance improvements. These conflicting findings underscore the lack of coherence in experimental outcomes and emphasize the ongoing challenge of finding a universally best AL strategy.

Given these challenges, it is natural to explore how an optimal selection strategy would look like. For this reason, oracle strategies (or oracle policies) have been introduced. \textbf{Oracle strategies}~\citep{zhou2021towards,sandrock2023exploring} aim to approximate the optimal selection in a feasible time by leveraging ground truth information typically unavailable to conventional AL selection strategies (e.g., access to all labels).  Despite their potential, oracle strategies remain underexplored in the literature, and existing methods often struggle with scalability issues.  \citet{zhou2021towards} introduced an oracle strategy that employs a simulated annealing search (SAS) to identify an optimal selection order given a fixed budget.  Even though they achieve impressive performance, the high number of search steps implies high computational cost, limiting an application to large datasets. \citet{sandrock2023exploring} introduced an iterative, non-myopic oracle strategy that selects instances based on both immediate and long-term performance improvements through a look-ahead approach. In their experiments, they mainly work with tabular data and employ a kernel-based classifier for fast retraining. Recently, \citet{werner2024a} proposed an oracle strategy as part of an AL benchmark, which we refer to as cross-domain oracle (CDO). Their approach greedily selects the instance, leading to the highest performance gain from a fixed number of randomly chosen instances. If no instance increases test performance, the selection is performed according to Margin. Due to selecting a single instance at a time, this approach requires retraining after each selection, significantly limiting its scalability for larger budgets.

Methodologically, our oracle strategy positions itself between the strategies of \citet{zhou2021towards} and \citet{werner2024a}. While \citet{zhou2021towards} optimize the selection over the entire labeled pool, and \citet{werner2024a} adopt a greedy, single-instance strategy, \methodname focuses explicitly on batch acquisitions. Consequently, our oracle strategy provides a less greedy perspective than \citep{werner2024a} by considering the collective performance improvement of instances within a batch. Additionally, our oracle searches more efficiently than \citep{zhou2021towards} by focusing on batch-level performance improvements instead of the entire labeled pool (cf.~\Cref{sec:method}). Searching for an optimal batch from a large pool is considerably simpler than searching for a (much larger) labeled pool. Consequently, \methodname demonstrates superior scalability, remaining effective even on large-scale datasets such as ImageNet.

\section{Notation}\label{sec:notation}
We consider classification tasks in a pool-based AL setting. Let $\vx \in \gX$ be an instance and $y \in \gY = \{1, \dots, K\}$ denote its corresponding label, where $K$ is the number of classes. Further, let $\gU \subset \gX$ be the large unlabeled pool and $\gL \subset \gX \times \gY$ be the labeled pool.  While $\gU$ is assumed to be sampled i.i.d.~from distribution $p(\vx)$, $\gL$ is typically biased towards instances a selection strategy considers informative.  Additionally, since our focus is on oracle strategies for evaluation, we consider an evaluation dataset $\gE$. At the start of AL, we initialize $\gL$  by randomly sampling $b$ instances from $\gU$. Then, we perform a total of $A$ AL cycles, selecting $b$ instances to label in each cycle. The total labeling budget is denoted by $B = b + A\cdot b$. As our model, we consider a DNN consisting of a feature extractor $h^\vphi: \gX \to \R^D$ and a classification head $g^\vtheta : \R^D \to \R^K$, where $\vphi$ and $\vtheta$ are trainable parameters. Hence, a DNN is a function $f^\vomega = (g^\vtheta \circ h^\vphi)(\vx)$ mapping an instance to the logit space, where $\vomega = \{\vphi, \vtheta\}$. The conditional distribution $p(y| \vx, \vomega) = [\softmax(f^\vomega(\vx))]_{y}$ is modeled through the output of the DNN.  We additionally consider the posterior distribution over parameters $p(\vomega|\gL)$ and the predictive distribution $p(y|\vx, \gL) = \E_{p(\vomega | \gL)}[p(y|\vx,\vomega)]$. 

\section{A Formalization of Performance-based Active Learning}\label{sec:method}
The \emph{main goal} of AL is to acquire the labeled pool that minimizes the model's error (or maximizes its performance) on unseen instances. We formalize the corresponding optimization problem by 
\begin{align}\label{eq:al_objective}
    \gL^\star = \argmin_{\gL\subset\gU}~\E_{p(\vx, y)}\big[\ell\big( y, p(y |\vx, \gL)\big)\big] \quad \text{subject to} \quad |\gL| = B,
\end{align}
where $\ell(y, p(y|\vx, \gL))$ denotes a loss function that quantifies the discrepancy between the true label $y$ and the probabilistic prediction $p(y |\vx, \gL)$.  Note the slight abuse of notation $\gL \subset \gU$ to signify that instances in $\gL$ are seen as a subset of those in $\gU$, even though $\gL$ includes labels and $\gU$ does not. Solving \cref{eq:al_objective} is computationally infeasible due to the enormous number of possible combinations of instances for $\gL$ and, more importantly, the absence of labels. Focusing on oracle strategies, we consider a supervised subset selection problem, i.e., the labels for all instances in $\gU$ are accessible to the oracle. 

While most AL selection strategies address the optimization problem in~\cref{eq:al_objective} indirectly (e.g., through uncertainty), some traditional strategies aim to optimize this objective directly~\citep{roy2001optimal}.  To this end, they employ a greedy approach, simplifying the problem of choosing $\gL$ to acquiring a single label per cycle. More specifically, for $B$ cycles, they select the instance $\vx_c$ for annotation that leads to the lowest error when added to the labeled pool:
\begin{align}\label{eq:single_instance_perf}
    \vx^\star = \argmin_{\vx_c \in\gU}~\E_{p(\vx, y)}\big[\ell\big( y, p(y |\vx, \gL^+)\big)\big],
\end{align}
where $\gL^+ = \gL \cup \{(\vx_c, y_c )\}$ is the extended labeled pool. This new optimization problem resolves the combinatorial problem by sequentially extending the labeled pool $\gL$.

However, the acquisition of a single instance per cycle poses several challenges when working with DNNs. The greedy selection in \cref{eq:single_instance_perf} only considers the immediate reduction in error rather than considering the long-term impact of instances~\citep{zhao2021uncertaintyaware}. This is particularly problematic for DNNs, where retraining with a single additional instance has little influence on the model's predictions~\citep{sener2018active}. Furthermore, retraining the model after each label acquisition is computationally impractical, especially in deep learning, where model training can take hours or even days~\citep{huseljic2025efficient}.

To address this problem, we reformulate the optimization problem in \cref{eq:single_instance_perf} to allow for batch selection. Specifically, over $\lceil \nicefrac{B}{b} \rceil$ cycles, we select a batch $\gB = \{\vx_{c_1}, \dots, \vx_{c_b}\}$ of $b$ instances that minimize the error according to
\begin{align}\label{eq:batch_perf}
    \gB^\star = \argmin_{\gB\subset\gU}~\E_{p(\vx, y)}\big[\ell\big( y, p(y |\vx, \gL^+)\big)\big],
\end{align}
where $\gL^+ = \gL \cup \{(\vx_i, y_i) \mid i = c_1, \dots, c_b \}$. Although this formulation introduces a combinatorial problem of selecting the batches $\gB$, it is simpler than the one in \cref{eq:al_objective}, as batch acquisition sizes are typically much smaller than the labeled pool in deep AL.\footnote{The complexity of this combinatorial problem depends on the acquisition size $b$. By assuming $1 < b \ll \tfrac{|\mathcal{U}|}{2}$, we obtain a problem that is less complex than the worst‐case with $\binom{|\mathcal{U}|}{|\mathcal{U}|/2}$ subsets.}

While evaluating all possible sets of $\gB$ remains infeasible, our idea is to effectively approximate the optimization problem by only considering a \emph{subset of the most promising batches}. The idea of directly selecting a batch from a set of batches is mostly avoided in deep AL~\cite{ash2021gone,hacohen2022active,gupte2024revisiting}. Typically, as heuristic strategies often yield batches with highly similar instances~\citep{kirsch2023stochastic}, the batch selection process is simplified through clustering of representations, emphasizing diversity by selecting informative instances from each cluster~\citep{ash2020deep,hacohen2022active,gupte2024revisiting}. This is especially important in early cycles of AL. However, it enforces diversity at every cycle, even when it is not beneficial~\citep{hacohen2022active}. In contrast, directly searching for the best batch, as done in \cref{eq:batch_perf}, allows the model itself to determine the most effective batch each cycle. While early cycles may benefit from diverse batches, later stages might favor more uncertain and less diverse ones. Additionally, directly considering promising batches rather than instances better captures the instances' long-term impact by evaluating how they collectively influence performance, leading to less greedy behavior.

\section{An Efficient Oracle Strategy for Deep Neural Networks}
We consider the objective in \cref{eq:batch_perf} to build an oracle strategy approximating optimal batch selection that can be efficiently applied in deep learning settings. Our proposed solution can be expressed as follows:
\def\fontscale{0.8}
\begin{align}\label{eq:components}
    \gB^\star = 
    \underbrace{{\argmin_{\gB\subset\gU}}}_{\text{\color{blue}\scalebox{\fontscale}{Batch Selection}}}
    ~\hspace*{-0pt}
    \underbrace{\vphantom{\argmin_{\gB\subset\gU}}\E_{p(\vx, y)}}_{\text{\color{teal}\scalebox{\fontscale}{Performance Estimation}}}
    \hspace*{-10pt}
    \big[\ell\big( y, \underbrace{\vphantom{\argmin_{\gB\subset\gU}}p(y |\vx, \gL^+)}_{\text{\color{purple}\scalebox{\fontscale}{Retraining}}}\big)\big] \equiv 
    {\color{blue}\argmin_{\gB\in \{\gB_1, \dots, \gB_T\}}} {\color{teal}\sum_{(\vx,y) \in \gE}
    \hspace*{0pt} \mathds{1}\big[ y \neq \argmax_{c \in \gY} {\color{purple}p(c |\vx, \gL^+)}\big]}
\end{align}
The optimization problem comprises three key components: \emph{Batch selection} involves identifying an optimal batch $\gB$ that yields the largest performance improvement, \emph{performance estimation} considers how to evaluate the model's performance when trained with additional data, including evaluation dataset $\gE$ and loss function~$\ell$, and \emph{retraining} refers to the process of efficiently retraining the DNN and computing updated predictions $p(y |\vx, \gL^+)$. In this section, we focus on how to efficiently implement each of these components.

\subsection{Batch Selection}\label{subsec:batch_selection}
As described in~\Cref{sec:method}, searching for the optimal batch introduces a combinatorial problem. For example, with an unlabeled pool of 1,000 instances and an acquisition size of 10, the number of possible batches is $\binom{|\gU|}{b} = \binom{1000}{10} = 2.63 \cdot 10^{23}$, making it computationally infeasible to iterate over all batches. In \methodname, we address this by restricting the search space to a fixed subset of $T \ll \binom{|\gU|}{B}$ candidate batches $\{\gB_1, \dots, \gB_T\}$. Consequently, the effectiveness of this approximation depends on the particular choice of those candidate batches. A naive approach is to solely draw batches randomly from the unlabeled pool
\begin{align}
    \gB_t \sim \operatorname{Unif}\left([\gU]^b\right),
    \label{eq:naive}
\end{align}
where $\operatorname{Unif}(\cdot)$ denotes uniform sampling and $[\gU]^b$ denotes all possible subsets of $\gU$ with size $b$. However, this approach might be inefficient because randomly selecting batches from the unlabeled pool ignores information about the data distribution or the model. In the example above, even if billions of near-optimal batches exist, the probability that a random sample of 100 candidate batches contains one of them is almost negligible.

For this reason, we suggest selecting a set of candidate batches through existing selection strategies. Recent studies~\citep{hacohen2022active,munjal2022towards,werner2024a} have shown that most strategies lack robustness across varying AL scenarios (e.g., a strategy effective for low budgets may not perform well for higher budgets). Given these insights, we leverage multiple perspectives of a variety of state-of-the-art selection strategies with complementary goals. By incorporating strategies that prioritize diversity or representativeness, we enhance exploration for lower budgets.  Similarly, emphasizing uncertainty or model change supports exploitation for higher budgets. Constructing candidate batches $\{\gB_1,\ldots, \gB_T\}$ in this manner, and then selecting the one that minimizes the error, naturally balances exploration and exploitation.  In principle, all strategies from the literature are potentially suitable for our oracle strategy. Furthermore, our oracle is highly flexible, since newly proposed strategies can be seamlessly integrated. Here, we focus on a carefully chosen set of state-of-the-art strategies (cf.~\Cref{tab:strategies}) that are selected based on three jointly considered key criteria:
\begin{itemize}
    \item \textbf{Coverage of relevant heuristics:} The selection strategies encompass all heuristics discussed in \Cref{sec:rel_work}.
    \item \textbf{State-of-the-art performance:} These selection strategies have consistently demonstrated strong performance in research.
    \item \textbf{Efficient computation:} Each strategy is associated with low computational costs to ensure scalability to large-scale datasets with many instances, classes, and/or feature dimensions.
\end{itemize}
Additionally, for two clustering-based strategies, we also include a supervised version that exploits labels to ensure each cluster corresponds to a class. We found that this is particularly valuable in tasks with suboptimal representations when clustering is difficult.

Preselecting candidate batches helps solving~\cref{eq:batch_perf} more effectively, yet the number of batches $T$ that can be considered for the search is constrained by the available \emph{deterministic} strategies.  Furthermore, despite their computational efficiency, some selection strategies still can become costly for large unlabeled pools. To address this, we propose to select multiple candidate batches by applying each strategy to randomly sampled candidate pools $\gC_1, \ldots, \gC_T \subset \gU$, each constrained by a maximum pool size $k_{\max}$. When choosing $k_{\max}$, we simply aim to ensure a representative subset of the unlabeled pool $\gU$. Beyond that, we found the choice to have little impact on performance and it can mainly be adjusted to improve selection speed. This reduces computational cost and memory requirements while allowing us to increase the number of candidate batches, even for deterministic selection strategies. Furthermore, we found it useful to vary the size of the candidate pools, as some strategies are prone to outliers or selecting similar instances (cf.~\cref{app:var_cand_pools}). The proposed algorithm is detailed in \Cref{alg:batch_selection}.
\begin{figure}[!ht]
\begin{minipage}[b]{.65\linewidth}
\begin{algorithm}[H]
\renewcommand{\algorithmicindent}{1em}
\small
\caption{Candidate Batch Selection}
\label{alg:batch_selection}
\begin{algorithmic}[1] 
    \Require Batch size $b$, number of batches $T$, selection strategies $\gS = \{s_1, \ldots, s_o\}$, maximum candidate pool size $k_{\max}$, unlabeled pool $\gU$, labeled pool $\gL$, model $\vomega$ 
        \State ${\gB}_{\text{cand}} \gets \emptyset$
        \For{each selection strategy $s \in S$} 
            \State $\hat{T} \gets \lfloor \nicefrac{T}{|S|} \rfloor$ \Comment{Determine the number of batches per strategy}
            \For{repeat $\hat{T}$ times} 
                \State $k \gets \operatorname{Unif}(\{b, \ldots, k_{\max}\})$ \Comment{Sample the size $k$ of the candidate pool}
                \State $\gC \gets \operatorname{Unif}([\gU]^{k})$ \Comment{Sample a candidate pool $\gC \subset \gU$ of size $k$}
                \State $\gB \gets s(\gC, \gL, b, \vomega)$ \Comment{Apply selection strategy $s$ on candidate pool $\gC$}
                \State $\gB_{\text{cand}} \gets \gB_{\text{cand}} \cup \{\gB\}$ \Comment{Extend $\gB_{\text{cand}}$ with batch $\gB$}
            \EndFor 
    \EndFor
    \State \Return $\gB_\text{cand}$
\end{algorithmic}
\end{algorithm}
\end{minipage} 
\hfill
\begin{minipage}[b]{.32\linewidth}
    \begin{table}[H]
    \setlength{\tabcolsep}{2pt}
    \scriptsize
    \centering
    \caption{Employed selection strategies for sampling candidate batches with their main characteristics. Strategies marked with * use labels as clusters.}\label{tab:strategies}
    \begin{tabular}{r c c c }
        \toprule
        \rowcolor{gray!30}
        \textbf{AL Strategy} & \textbf{Unc} & \textbf{Repr}& \textbf{Div} \\
        \midrule
        Random~\citeyearpar{settles2009active}    & \xmark & \cmark & \cmark  \\
        Margin~\citeyearpar{settles2009active}    & \cmark & \xmark & \xmark  \\
        CoreSets~\citeyearpar{sener2018active} & \xmark & \xmark & \cmark  \\
        BADGE~\citeyearpar{ash2020deep}     & \cmark & \xmark & \cmark  \\
        FastBAIT~\citeyearpar{huseljic2024fast}      & \cmark & \cmark & \cmark  \\
        TypiClust~\citeyearpar{hacohen2022active} & \xmark & \cmark & \cmark  \\
        AlfaMix~\citeyearpar{parvaneh2022active}    & \cmark & \cmark & \cmark  \\
        DropQuery~\citeyearpar{gupte2024revisiting} & \cmark & \cmark & \cmark  \\
        TypiClust*~\citeyearpar{hacohen2022active} & \cmark & \cmark & \cmark  \\
        DropQuery*~\citeyearpar{gupte2024revisiting} & \cmark & \cmark & \cmark  \\
        \bottomrule
    \end{tabular}
    \end{table}
\end{minipage}
\end{figure}

\subsection{Performance Estimation}
Evaluating the model performance after retraining with every candidate batch is essential to determine how much the model has improved. In a supervised setting, this evaluation is typically performed using a labeled validation dataset. However, in AL, such labeled validation datasets are typically not available. Consequently, performance estimation becomes an unsupervised problem and requires alternative methods to assess the model's effectiveness. Performance-based selection strategies, such as expected error reduction~\citep{roy2001optimal}, address this challenge by estimating the \emph{expected error} that considers the factorization of the joint distribution $p(\vx, y)$. 

For \methodname, we aim to establish an oracle strategy, i.e., approximating the best possible strategy that leverages all available information. Thus, it is justified to utilize the test split of a given dataset as our evaluation dataset $\gE$ to estimate model performance. This ensures that the performance of the retrained model is accurately captured, and that the selected batches indeed result in the highest gain. Additionally, for the loss function $\ell(\cdot)$, the zero-one loss works best. This is because AL strategies are typically evaluated via accuracy learning curves and the zero-one loss directly corresponds to the accuracy. Additionally, our experiments in \Cref{sec:ablations} show that the Brier score also works well. This is likely due to being a proper scoring rule, leading to a fine-grained assessment of probabilistic predictions~\citep{ovadia2019can}. 

\subsection{Retraining}
Retraining, particularly with DNNs, is the most time-consuming step in performance-based AL. Generally, batch selection is employed to avoid frequent training after each AL cycle. In our oracle strategy, however, the DNN is to be retrained for each candidate batch, resulting in $T$ retraining repetitions per selection. Although this is faster than retraining after a single instance, the computational overhead is still considerable and limits the size and the number of candidate batches that can be evaluated. For larger-scale datasets such as ImageNet, this process gets increasingly expensive as the labeled pool $\gL$ grows, making naive retraining with each candidate batch computationally infeasible. Moreover, retraining must accurately reflect changes in $\gL$ to capture which batches truly improve performance. Specifically, even small changes in $\gL$ can considerably alter the training dynamics of large DNNs (e.g., change of optimal hyperparameters) potentially yielding noisy and unreliable performance estimates.

For this reason, we propose a selection-via-proxy approach~\citep{coleman2020selection} that decouples the retraining process during the selection from the usual cyclic training in AL. Specifically, we freeze the feature extractor’s parameters $\vphi$ and only retrain the final linear layer $\vtheta$. This not only significantly reduces retraining time but also enhances stability by restricting parameter updates to a much simpler model. Furthermore, to assess the candidate batches during the selection, we reduce the number of retraining epochs from 200 (as used in our experiments \emph{after} selection) to 50. As shown in~\Cref{sec:ablations}, this is sufficient to identify influential candidate batches while reducing computation substantially.

While this approach enables the efficient use of \methodname, there are additional approaches to improve retraining efficiency. For instance, by employing continual learning strategies~\citep{huseljic2025efficient}, the retraining time of the DNN scales only with the new batch $\gB$, making the process largely independent of the size of the extended dataset $\gL^+$. As these approaches involve new training hyperparameters, we opt for the simpler variant of retraining only the last layer and leave the exploration of more complex alternatives for future work. 

\section{Comparison of Time Complexity}
We investigate \methodname's time complexity of selecting a batch in comparison to existing oracle strategies. Specifically, we consider SAS~\citep{zhou2021towards} and the recently introduced CDO~\citep{werner2024a}. The time complexities in $\gO$-notation are summarized in \Cref{tab:o_notation}, where $\texttt{train-eval}(\vtheta, \gL, \gE)$ denotes the cost of retraining model $\vtheta$ on dataset $\gL$ and then evaluating it on dataset $\gE$. Since all oracle strategies primarily differ in terms of retraining and evaluation frequency, we also report the hyperparameters recommended by the respective approaches along with the total number of training instances processed during selection.

CDO~\citep{werner2024a} greedily selects the instance with the highest performance improvement from $m$ randomly sampled instances. This requires $b \cdot m$ retrainings for a batch of size $b$. Due to its greedy nature, CDO acquires instances sequentially and retrains each time on a labeled pool expanded by one instance, denoted as $\gL^{+i}$.
SAS~\citep{zhou2021towards} performs simulated annealing and greedy refinement search steps, represented by parameters $s$ and $g$, respectively. Their approach requires $s + g$ retrainings, whereby the labeled pool $\gL^{+b}$ has been extended by a batch of $b$ instances. Additionally, as SAS evaluates the entire learning curve at each search step (rather than the improvement of a batch), retraining and evaluation times are multiplied by the total number of AL cycles $A$. 
In contrast to these strategies, \methodname depends solely on the number of candidate batches $T$, determined by the number of batches per strategy $s \in \gS$. Consequently, the retraining frequency remains independent of batch size $b$ and the number of cycles $A$, offering a significant advantage in terms of scalability.

For CDO, \citet{werner2024a} recommend setting $m = 20$, resulting in $20$ retrainings per instance selection within a batch. This quickly becomes infeasible with larger batch sizes that are common for more complex datasets requiring higher budgets. For example, a batch size of $b = 100$ would require to retrain $2{,}000$ times, which becomes especially expensive towards the end of the AL process, as the labeled pool $\gL$ increases in size.
Similarly, SAS recommends $s = 25{,}000$ simulated annealing steps and $g = 5{,}000$ greedy refinement steps. However, these parameters determine the frequency of retrainings to obtain the final optimized pool, i.e., $|\gL| = B$. As we compare the frequency of retraining per batch, we divide these values by the total number of AL cycles used in our experiments ($A = 20$). This results in $20 \cdot 1{,}500$ retrainings, which is even less scalable to larger datasets. Finally, considering the number of processed training instances, CDO scales quadratically with batch size, posing a major bottleneck. For instance, with $b = 50$ and $|\gL| = 50$, CDO processes approximately 75k instances. In contrast, SAS and \methodname scale linearly, with \methodname achieving a substantially more efficient search, requiring only 11k instances in the same setting.

While our approach involves less frequent retraining, it additionally requires preselecting candidate batches using specific selection strategies. This step introduces extra computational overhead for batch selection. However, with the set of efficient selection strategies we proposed in~\Cref{tab:strategies}, combined with sampling candidate pools significantly smaller than the entire unlabeled pool, this computational burden remains negligible (cf.~\Cref{sec:experiments}). Nevertheless, it is important to highlight that when extending \methodname with more contemporary selection strategies, one must ensure that the computational cost associated with these strategies is considered. In our experiments, we analyzed how the number of candidate batches per strategy affects performance and, in general, increasing this number will always lead to an improvement. However, in \Cref{sec:ablations}, we found that increasing the number of candidate batches beyond 10 did not yield notable benefits. Thus, we adopt 10 batches per strategy, resulting in a total of $T=100$ candidate batches. 
\begin{table}[!t]
    \centering
    \scriptsize
    \caption{Summary of time complexities of oracle strategies.}
    \setlength{\tabcolsep}{11pt}
    \begin{tabular}{lrrr}
         \toprule
         \rowcolor{gray!15}
          &
         \multicolumn{1}{c}{{\textbf{Time Complexity}}} &
         \multicolumn{1}{c}{\textbf{Recommended}} &
         \multicolumn{1}{c}{\textbf{\# Processed Training }} \\
         \rowcolor{gray!15}
         \multicolumn{1}{c}{\multirow{-2}{*}{\textbf{Oracle Strategy}}} & \multicolumn{1}{c}{{\textbf{per Batch}}} & \multicolumn{1}{c}{{\textbf{Hyperparameters}}} & \multicolumn{1}{c}{\textbf{Instances }}\\
         
         \midrule
         CDO~\citep{werner2024a} & $\gO( m \cdot \sum_{i=1}^b\texttt{train-eval}(\vtheta, \gL^{+ i}, \gE))$ 
         & $m = 20$
         & $20 \cdot (b \cdot |\gL| + \frac{b(b + 1)}{2})$ \\
         SAS~\citep{zhou2021towards} & $\gO( (s + g) \cdot A \cdot \texttt{train-eval}(\vtheta, \gL^{+b}, \gE))$ & $s = 1{,}250, g = 250$ 
         & $1{,}500 \cdot A \cdot (|\gL| + b)$ \\
         \methodname & $\gO(T \cdot \texttt{train-eval}(\vtheta, \gL^{+ b}, \gE))$ & $T = 100$ 
         & $10 \cdot |\gS| \cdot (|\gL| + b)$ \\
         \bottomrule
    \end{tabular}
    \label{tab:o_notation}
\end{table}

\section{Empirical Evaluation of \methodname: Oracle-Level and State-of-the-art AL Comparisons}\label{sec:experiments}
We evaluate our oracle strategy for the task of image classification. After detailing the experimental setup, we begin with a comparison of \methodname to other oracle strategies. Afterward, we benchmark our approach against state-of-the-art selection strategies across ten image datasets. Our evaluation is driven by four research questions:
 \begin{itemize}
     \item[\hypertarget{rq1}{RQ\textsubscript{1}}:] \emph{Given comparable computational resources, can \methodname match or exceed the accuracy improvements of state-of-the-art deep AL oracle strategies (CDO, SAS)?}
     \item[\hypertarget{rq2}{RQ\textsubscript{2}}:] \emph{Does \methodname consistently match or surpass the highest test accuracy that any current state-of-the-art AL strategy achieves at every cycle, making it a practical oracle strategy?}
     \item[\hypertarget{rq3}{RQ\textsubscript{3}}:] \emph{Where lies the greatest potential for improving state-of-the-art AL strategies when comparing them to \methodname across cycles, datasets of varying complexity, and different models?}
     \item[\hypertarget{rq4}{RQ\textsubscript{4}}:] \emph{What insights regarding AL research can we get by analyzing which selection strategy's candidate batch has been chosen by \methodname?}
 \end{itemize}
In a nutshell, we find that \methodname not only ties or outperforms CDO/SAS in most settings (RQ\textsubscript{1}) but also serves as a reliable oracle strategy (RQ\textsubscript{2}), with the biggest performance improvements appearing in large-scale multiclass s settings (RQ\textsubscript{3}). Moreover, our results highlight that each AL strategy contributes to the selection of \methodname and that there is no single best strategy across datasets or cycles within a dataset (RQ\textsubscript{4}), emphasizing potential advantages in using an ensemble-based AL approaches that combine multiple strategies~\citep{donmez2007dual}.
  
\subsection{Experimental Setup}
We evaluate our oracle strategy on ten image datasets of varying complexity. For each dataset, we conduct 20 AL cycles, starting with a randomly selected initial pool of $b$ instances, and selecting an additional batch of $b$ new instances in each subsequent cycle. Batch sizes were determined by analyzing the convergence of learning curves obtained via Random sampling. Consequently, the complexity of each dataset is indicated not only by the number of classes $K$ but also by the respective batch size. \Cref{tab:datasets} summarizes these datasets, their number of classes $K$, and the employed AL batch sizes $b$.

We employ two pretrained Vision Transformers (ViTs)~\citep{dosovitskiy2020image} that are complemented by a randomly initialized fully connected layer. Specifically, we use DINOv2-ViT-S/14~\citep{oquab2024dinov2} (22M parameters) and SwinV2-B~\citep{liu2022swin} (88M parameters), whose final hidden layers provide feature dimensions of $D = 384$  and $D = 1024$, respectively.
 \begin{wraptable}{r}{.4\linewidth}
    \scriptsize
    \centering
    \caption{Overview of datasets, showing number of classes $K$ and batch size $b$.}
    \label{tab:datasets}
    \setlength{\tabcolsep}{2pt}
    \begin{tabular}{lrr}
    \toprule
    \rowcolor{gray!30}
    \textbf{Dataset} & \textbf{\# Classes} ($K$) & \textbf{Batch Size ($b$)} \\
    \midrule
    \makebox[65pt][l]{CIFAR-10~\citeyearpar{krizhevsky2009learning}}   & 10    & 10 \\
    \makebox[65pt][l]{STL-10~\citeyearpar{coates2011analysis}}         & 10    & 10 \\
    \makebox[65pt][l]{Snacks~\citeyearpar{snacks2023dataset}}          & 20    & 20 \\
    \makebox[65pt][l]{Flowers102~\citeyearpar{nilsback2008flowers}}    & 102   & 25 \\
    \makebox[65pt][l]{Dopanim~\citeyearpar{herde2024dopanim}}          & 15    & 50 \\
    \makebox[65pt][l]{DTD~\citeyearpar{cimpoi14describing}}            & 47    & 50 \\
    \makebox[65pt][l]{CIFAR-100~\citeyearpar{krizhevsky2009learning}}  & 100   & 100 \\
    \makebox[65pt][l]{Food101~\citeyearpar{bossard14}}                 & 101   & 100 \\
    \makebox[65pt][l]{Tiny ImageNet~\citeyearpar{le2015tiny}}          & 200   & 200 \\
    \makebox[65pt][l]{ImageNet~\citeyearpar{russakovsky2015imagenet}}  & 1000  & 1000 \\
    \bottomrule
    \end{tabular}
    \vspace{-10pt}
\end{wraptable}
The two differ both in size and in training paradigm: The former was trained via self-supervised learning, while the latter was pretrained on ImageNet in a supervised manner. Note that the ImageNet results obtained with the SwinV2-B backbone are not fully representative, as the model was pretrained on the same dataset. Nevertheless, we include them for completeness. After a batch is selected, each DNN is trained by fine-tuning the last layer on frozen representations for 200 epochs, employing SGD with a training batch size of 64, a learning rate of 0.01, and weight decay of 0.0001. In addition, we utilize a cosine annealing learning rate scheduler. These hyperparameters were determined empirically across datasets by investigating the loss convergence on validation splits. Note that the number of epochs here applies to training after an AL cycle once a batch is selected. In contrast, the \emph{retraining} epochs described in \Cref{sec:method} refer to those we use to assess candidate batches.

To evaluate the AL process, we examine the resulting learning curves of oracle and selection strategies. These include \emph{relative learning curves}, which represent the accuracy difference of each strategy compared to Random sampling, and the \emph{area under the absolute learning curves} (AULC). The corresponding absolute learning curves can be found in Appendix~\ref{app:lcs}. All reported scores are averaged over ten trials. For visual clarity, standard errors have been omitted from the figures. All benchmark experiments were conducted on servers equipped with NVIDIA Tesla V100 and A100 GPUs as well as AMD EPYC 7742 CPUs. Experiments involving runtime measurements were performed on a workstation with an NVIDIA RTX 4090 GPU and an AMD Ryzen 9 7950X CPU in a controlled environment to ensure reproducibility and minimize external influences on the measurements (e.g., by ensuring consistent CPU load and hardware configuration).

\subsection{Benchmark Results}
To answer \hyperlink{rq1}{RQ\textsubscript{1}}, we first align hyperparameters of CDO and SAS to closely match the empirical runtime of \methodname, and then compare the resulting learning curves. In principle, with a longer runtime we consider more combinations to solve the combinatorial problem, inevitably improving each oracle’s performance. Accordingly, we ensure a fair comparison by approximately equalizing runtimes.  Due to the high computational effort of oracle strategies, we focus on four datasets using the DINOv2-ViT-S/14 model. When aligning hyperparameters, we made sure that CDO and SAS have at least as much compute as \methodname, ensuring that any performance advantage is not due to differing computational resources. Importantly, all oracle strategies use the same retraining procedure, ensuring differences in performance are solely due to the selection mechanism. The employed hyperparameter settings are summarized in \Cref{tab:hyperparams_oracles}, while the associated empirical runtimes can be found in \Cref{tab:runtimes}. Note that different configurations of $\gS$ can yield varying runtimes and outcomes. Rather than selecting the ensemble of strategies to optimize runtime, we include all strategies from~\Cref{tab:strategies} to prioritize robustness across settings (e.g., datasets and models).
\begin{table}[!ht]
    \centering
    \caption{Hyperparameters of oracle strategies under runtime constraints equivalent to \methodname.}
    \scriptsize
    \setlength{\tabcolsep}{10pt}
    \label{tab:hyperparams_oracles}
    \begin{tabular}{l|r|rrrr}
        \toprule
        \rowcolor{gray!30}
        \textbf{Oracle} & \textbf{Default} & \textbf{CIFAR-10} ($b=10$) & \textbf{Snacks} ($b=20$) & \textbf{Dopanim} ($b=50$) & \textbf{DTD} ($b=50$) \\
        \midrule
        \methodname & $T=100$ & $T=100$ & $T=100$ & $T=100$ & $T=100$ \\
        CDO         & $m=20$ & $m=20$ & $m=10$ & $m = 4$ & $m = 3$ \\
        SAS         & $s=25{,}000, g=5{,}000$ & $s=250, g=10$ & $s=225, g=10$ & $s=215, g=10$ & $s=150, g=10$\\
        \bottomrule
    \end{tabular}
\end{table}

\begin{figure}[!ht]
    \centering
    \includegraphics[width=\linewidth]{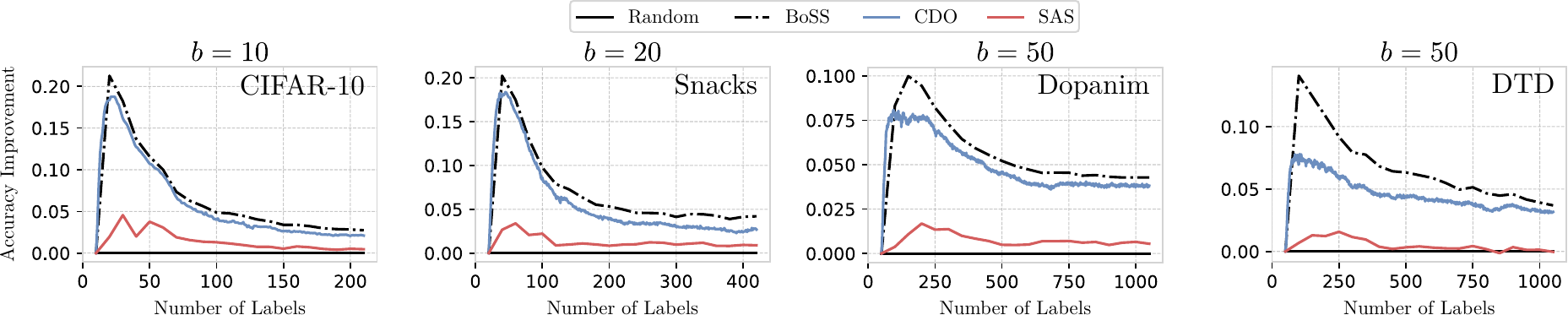}
    \caption{Relative learning curves of oracle strategies with aligned runtimes using DINOv2-ViT-S/14.}\label{fig:oracle_experiments}
\end{figure}

\Cref{fig:oracle_experiments} presents the learning curves reporting the relative accuracy improvement of each oracle strategy over Random sampling. Absolute learning curves together with learning curves that report the performance using default hyperparameters can be found in \Cref{app:lcs}.
All oracle strategies outperform random selection in terms of accuracy. 
However, unlike the other oracles, SAS yields only marginal accuracy improvements as the number of search steps were reduced considerably in comparison to the authors' recommendation. 
\begin{wraptable}{r}{0.45\linewidth}
\scriptsize
\centering
\setlength{\tabcolsep}{3pt}
\caption{Empirical runtimes of oracle strategies with adapted and default hyperparameters.}\label{tab:runtimes}
\begin{tabular}{rrrrr}
\toprule
\rowcolor{gray!30}
\textbf{Oracle} & \textbf{Cifar-10} & \textbf{Snacks} & \textbf{Dopanim} & \textbf{DTD} \\
\midrule
\methodname    & 10:07   & 13:19  & 30:47   & 22:07   \\
\midrule
CDO (Adapted) & 10:26   & 14:11  & 33:25   & 24:56   \\
SAS (Adapted) & 10:20   & 13:24  & 31:56   & 22:14   \\
\midrule
CDO (Default)  & 10:26   & 28:22  & 2:47:08 & 2:46:14 \\
SAS (Default)  & 17:22:39 & 24:19:42 & 61:52:42 & 62:34:13 \\
\bottomrule
\end{tabular}
\end{wraptable}
Restoring the recommended number improves accuracy but at the cost of much higher computation. 
In contrast, CDO and \methodname are much more effective, achieving approximately up to 20\% accuracy improvement on CIFAR-10 and Snacks, and around 10\% on Dopanim and DTD.
We see that especially in larger scale settings with higher batch sizes, \methodname outperforms the other oracle strategies. As shown by the hyperparameters of CDO in \Cref{tab:hyperparams_oracles}, increasing the AL batch size ($b = 50$) results in a considerable reduction of its hyperparameter ($m = 4$ and $m=3$). Since $m$ denotes the number of randomly sampled instances from which CDO selects the best, further increasing the batch size would prevent aligning its runtime to that of \methodname, highlighting CDO’s inefficiency with larger batch sizes. \emph{Overall, \methodname consistently matches or surpasses the performance of other competing oracle strategies across all datasets under comparable computational resources.}

\begin{figure}[!b]
    \centering
    \subfloat[DINOv2-ViT-S/14]{\includegraphics[width=\linewidth]{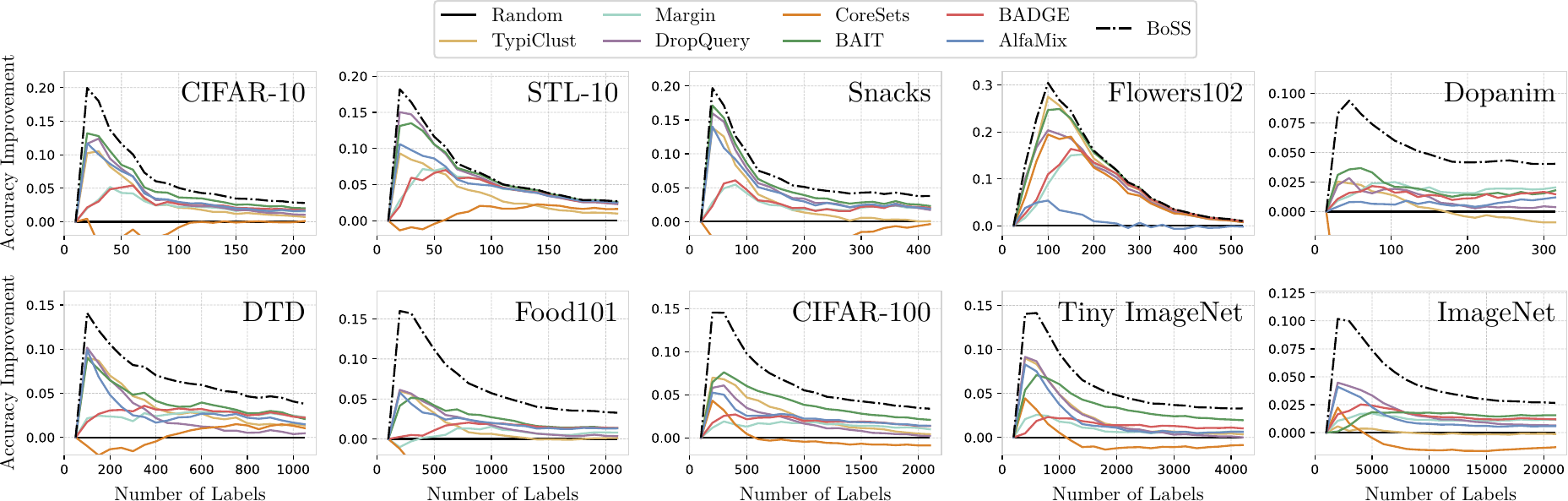}}
    \vspace{1em}
    \subfloat[SwinV2-B]{\includegraphics[width=\linewidth]{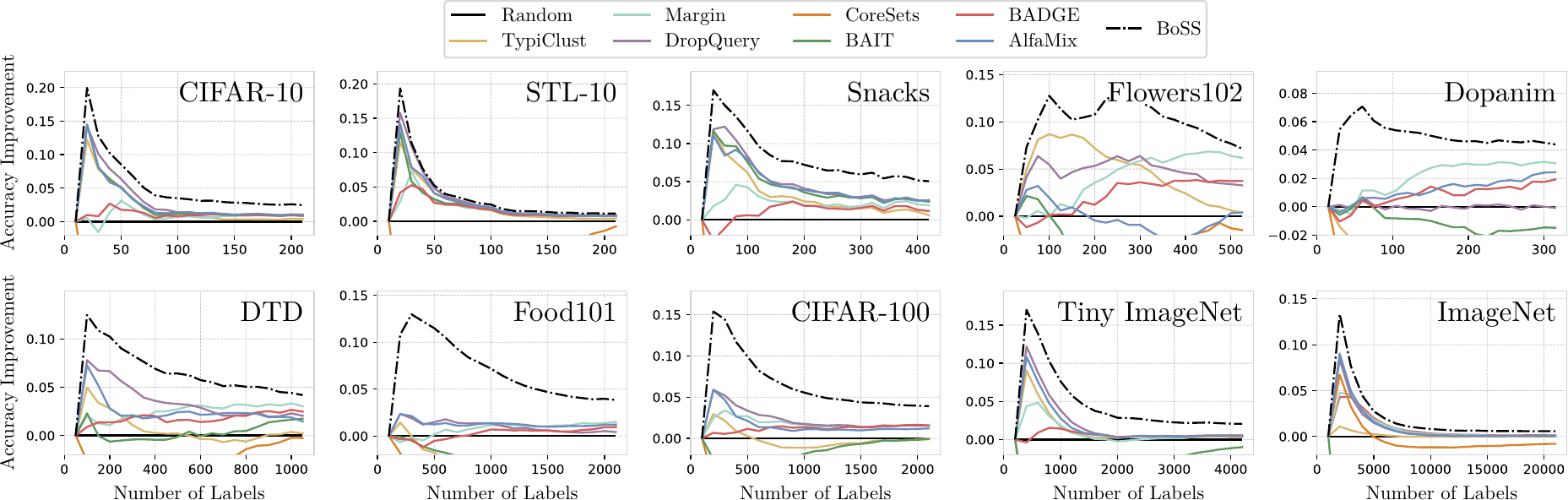}}
    \caption{Relative learning curves achieved by \methodname and state-of-the-art selection strategies at each annotation cycle for different pretrained models.
    }
    \label{fig:sota_experiments}
\end{figure}
To answer \hyperlink{rq2}{RQ\textsubscript{2}}, we consider the learning curves in \Cref{fig:sota_experiments}, depicting relative accuracy improvements over Random sampling. They clearly demonstrate that \methodname consistently outperforms existing AL strategies across all cycles and datasets, for both DINOv2 and SwinV2. \emph{Consequently, we assume \methodname to be a reliable oracle strategy for deep AL selection strategies.} Notably, while the overall accuracy improvement provided by \methodname across the entire AL cycle is modest for simpler datasets such as CIFAR-10 and STL-10, it becomes substantially more pronounced with more challenging datasets, particularly those with more than 20 classes. To complement these findings, we further provide a detailed budget-regime analysis in Appendix~\ref{app:budget_analysis} and additional experiments on text data in Appendix~\ref{app:text_results}.

To answer \hyperlink{rq3}{RQ\textsubscript{3}}, we compare \methodname to the best-performing AL strategies per dataset. In \cref{fig:sota_experiments}, we see substantial accuracy differences across all stages of AL.
Taking ImageNet with DINOv2 as an example, \methodname achieves approximately twice the accuracy improvement compared to the best-performing AL strategy.
The significant gap during the initial exploration phase suggests that there is potentially still room to address the cold-start problem more effectively. Similarly, the performance gap at later cycles indicate potential shortcomings during the exploitation phase, suggesting that current state-of-the-art AL strategies may struggle to effectively identify instances most valuable for further refinement.
Comparisons across datasets reveal that the potential for improving AL strategies correlates with dataset complexity. For example, on less complex datasets such as CIFAR-10, STL-10, or Snacks, AL strategies generally perform closer to the oracle. Conversely, more challenging datasets such as Food101, CIFAR-100, Tiny ImageNet, and ImageNet exhibit substantial gaps. \emph{This finding indicates that large-scale multiclass settings may be a particularly relevant area for further study of AL strategies.}
Finally, examining AL strategies across different models demonstrates variability in their effectiveness. For instance, strategies that perform closely to the oracle with DINOv2 on datasets like Snacks or Flowers102 exhibit notably larger performance gaps when used with SwinV2. 
\emph{This discrepancy suggests that further work on more robust and model-agnostic AL strategies could be beneficial for achieving consistent performance across diverse models.}
Alternatively, any newly proposed strategy should include detailed analyses of its failure cases (e.g., being limited to a specific architecture), enabling practitioners to understand the specific scenarios in which it may underperform. 

\begin{figure}[!ht]
    \centering
    \includegraphics[width=\linewidth]{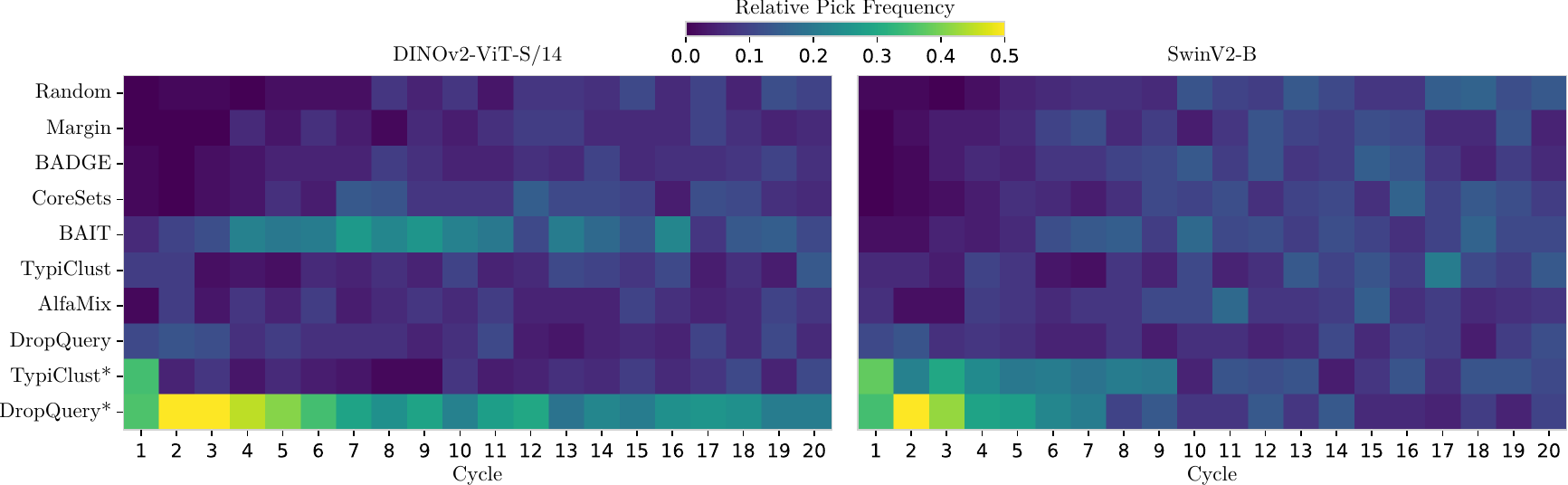}
    \caption{Average relative pick frequency of AL strategies by \methodname across cycles averaged over all datasets.}
    \label{fig:avg_picks}
\end{figure}
Finally, to answer \hyperlink{rq4}{RQ\textsubscript{4}}, we examine which AL strategy was selected by \methodname across cycles. For this purpose, \cref{fig:avg_picks} depicts the relative pick frequency of the strategies for both models, each averaged over all datasets. We notice that each strategy is considered at a certain point in the AL cycle, which indicates their respective contribution to the overall performance of \methodname. However, especially at the beginning of AL, the strategies DropQuery* and TypiClust* with supervised cluster assignments dominate, underscoring the importance of representativeness instances at that stage. Considering the more detailed dataset-specific pick frequencies in \Cref{app:picks_per_dataset} reveals that this effect is particularly pronounced for large-scale multiclass settings. This suggests that more sophisticated AL strategies may be required in these scenarios and that the supervised selection strategies play an important role for the performance of \methodname. Nevertheless, other strategies such as BAIT are also selected regularly. Notably, towards the end of the AL process, as we approach the convergence region of the learning curves, Random sampling is increasingly taken into account. This suggests that in certain stages none of the specialized AL strategies provide effective candidate batches, which is why more investigation of strategies for exploitation could be beneficial.
Similarly, the tendency towards the supervised selection strategies such as DropQuery in the beginning of AL indicates that current strategies lack in identifying effective batches for exploration.
\emph{However, most importantly, no single AL strategy consistently outperforms others across all phases of AL. This indicates that the best strategy at a given cycle can vary significantly depending on the context and stage of the AL process.} Consequently, we believe that advancing AL research requires a stronger focus on ensemble-based AL strategies~\citep{donmez2007dual,hacohen2023how}. These strategies integrate multiple strategies and adaptively select the most suitable strategy for the current context, thereby leveraging the specific strengths of each individual strategy.

\section{Analytical Evaluation: Ablations and Sensitivity Analyses}\label{sec:ablations}
To better understand the contributions of each component in~\cref{eq:components}, we conduct a series of ablation studies and experiments on a representative subset of datasets while fixing the backbone to DINOv2-ViT-S/14. These analyses aim to isolate the impact of each design choice, evaluate robustness under different conditions, and provide deeper insight into the mechanisms that determine performance.
 
\subsection{Selection of Candidate Batches}
\begin{wrapfigure}{r}{.45\linewidth}
    \vspace{5pt}
    \centering
    \includegraphics[width=\linewidth]{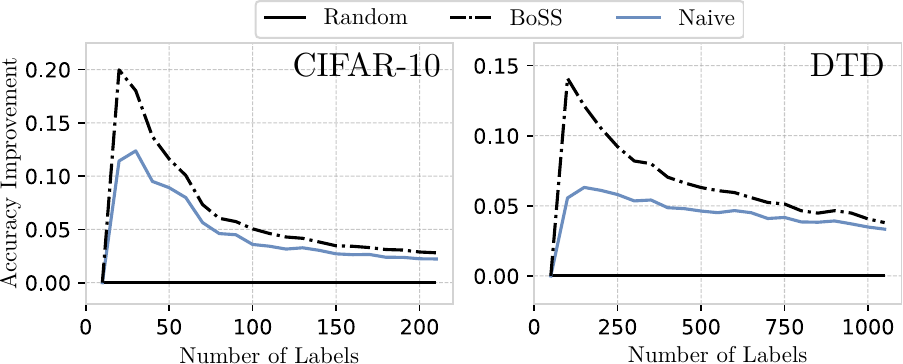}
    \caption{Relative learning curves of \methodname with naively selected candidate batches vs.~our algorithm.}\label{fig:naive_vs_algo}
    \vspace{-15pt}
\end{wrapfigure}

We compare the proposed selection of candidate batches from \Cref{alg:batch_selection} against the naive selection from \cref{eq:naive}. Unlike Random sampling, which directly selects $\gB^\star$ at random, the naive selection generates the candidate batches randomly. Relative learning curves for CIFAR-10 and DTD are shown in \Cref{fig:naive_vs_algo}. While the naive selection of candidate batches leads to a better performance than Random sampling, \Cref{alg:batch_selection} considerably improves performance, indicating the importance of a proper candidate batch selection.  Moreover, as shown in \Cref{tab:num_batches}, increasing the number of candidate batches $T$ yields further performance gains. Since improvements beyond 10 batches per strategy were negligible, we opted for this value in our experiments, resulting in a total of $T=100$ candidate batches.
Finally, varying the size of candidate pools leads to further improvements, as demonstrated in \Cref{app:var_cand_pools}.

Additionally, we examine influence of selection strategies $\mathcal{S}$. 
For this, we first applied \methodname with all selection strategies from~\Cref{tab:strategies} on CIFAR-10, analyzing the frequency with which each strategy's candidate batch was selected. This analysis enables us to identify the most influential strategies specifically for CIFAR-10. Subsequently, we iteratively applied \methodname to the Dopanim dataset, progressively incorporating the next most influential strategy according to the order established earlier. 
This way, we avoid data-specific overfitting since, in reality, the optimal order for a given task is unknown before running the experiments. Throughout this process, we maintain a constant total candidate batch size $T$. Consequently, the addition of each new strategy proportionally decreased the number of candidate batches generated by previously included strategies.
As shown in~\Cref{tab:strategy_ablation}, this sequential inclusion of strategies consistently improves the resulting AULC. These findings suggest that integrating more strategies generally enhances overall performance, and thus, we opt to include a variety of strategies when using \methodname. Intuitively, the more selection strategies we incorporate, the better the robustness of \methodname should be across models, datasets, and domains.
Importantly, including additional strategies does not degrade performance, suggesting that \methodname is robust to weaker strategies in the ensemble.

Finally, we examine the impact of batch size $b$, which directly affects the search space of potential candidate batches. As $b$ increases, the number of possible subsets generated from $\gU$ grows, making the combinatorial problem in~\cref{eq:batch_perf} more difficult. To investigate this impact, we run \methodname with different batch sizes $b$ on the same datasets multiple times. The AULCs in~\Cref{tab:impact_batch_size} on CIFAR-10 demonstrate that increasing the batch size by a factor of four ($4b = 40$) results in a marginal performance decrease. Similarly, for DTD, this batch size ($4b = 200$) yields a slightly more noticeable decline from 64.83 to 63.19. These findings highlight that the effectiveness of solving the combinatorial problem and identifying optimal candidate batches remains sensitive to the chosen batch size $b$. Nonetheless, the results presented in~\Cref{sec:experiments} show a substantial improvement over all considered state-of-the-art strategies. To further enhance \methodname with even larger batch sizes, increasing the number of candidate batches $T$ seems to be an effective strategy to mitigate potential performance losses.

\begin{table}[!t]
    \centering
    \scriptsize
    \setlength{\tabcolsep}{3pt}
    
    \begin{minipage}{.33\linewidth}
    \centering
    \captionof{table}{AULC of \methodname with varying number of batches per strategy.}\label{tab:num_batches}
    \begin{tabular}{rrr}
    \toprule
    \rowcolor{gray!30}
    \textbf{Batches per} &  &  \\
    \rowcolor{gray!30}
    \textbf{Strategy} & \multicolumn{1}{c}{\multirow{-2}{*}{\textbf{CIFAR-10}}} & \multicolumn{1}{c}{\multirow{-2}{*}{\textbf{DTD}}} \\
    \midrule
    \textbf{1}     & 89.90\textsubscript{$\pm$0.11} & 70.45\textsubscript{$\pm$0.12} \\
    \textbf{5}     & 90.45\textsubscript{$\pm$0.10} & 71.55\textsubscript{$\pm$0.14} \\
    \textbf{10}    & 90.70\textsubscript{$\pm$0.11} & 71.79\textsubscript{$\pm$0.11} \\
    \textbf{20}    & 90.83\textsubscript{$\pm$0.15} & 71.91\textsubscript{$\pm$0.15} \\
    \bottomrule
    \end{tabular} %
    \vspace{10pt}
    \caption{AULC of \methodname for different batch sizes.}\label{tab:impact_batch_size}
    \begin{tabular}{rcc}
        \toprule
        \rowcolor{gray!30}
        \textbf{Batch Size} & \textbf{CIFAR-10} & \textbf{DTD} \\
        \midrule
        $0.5\cdot b$        & 85.71\textsubscript{$\pm$0.32} & 68.41\textsubscript{$\pm$0.14} \\
        $b$                 & 85.62\textsubscript{$\pm$0.30} & 67.95\textsubscript{$\pm$0.16} \\
        $2\cdot b$          & 85.36\textsubscript{$\pm$0.31} & 67.37\textsubscript{$\pm$0.16} \\
        $4\cdot b$          & 84.95\textsubscript{$\pm$0.30} & 66.82\textsubscript{$\pm$0.12} \\
        \bottomrule
    \end{tabular}
    \end{minipage} %
    \begin{minipage}{.25\linewidth}
    \centering
    \vspace{43pt}
    \caption{AULC on Dop\-a\-nim when incorporating additional AL strategies.}\label{tab:strategy_ablation}
    \begin{tabular}{lc}
    \toprule
    \rowcolor{gray!30}
    \textbf{Strategies} & \textbf{AULC} \\
    \midrule
    Random & 75.24\textsubscript{$\pm$0.15} \\
    +DropQuery & 75.82\textsubscript{$\pm$0.17} \\
    +AlfaMix & 76.01\textsubscript{$\pm$0.16} \\
    +TypiClust & 76.08\textsubscript{$\pm$0.17} \\
    +BAIT & 76.28\textsubscript{$\pm$0.16} \\
    +CoreSet & 76.29\textsubscript{$\pm$0.20} \\
    +Margin & 76.26\textsubscript{$\pm$0.18} \\
    +BADGE & 76.35\textsubscript{$\pm$0.18} \\
    +DropQuery* & 76.48\textsubscript{$\pm$0.20} \\
    +TypiClust* & 76.52\textsubscript{$\pm$0.18} \\    \bottomrule
    \end{tabular}
    \end{minipage} %
    \begin{minipage}{.37\linewidth}
    \centering
    \caption{AULC of \methodname using different loss functions.}\label{tab:perf_estimation}
    \scriptsize
    \setlength{\tabcolsep}{6pt}
    \begin{tabular}{rcc}
    \toprule
    \rowcolor{gray!30}
    \textbf{Loss} ($\ell$) & \textbf{CIFAR-10} & \textbf{DTD} \\
    \midrule
    Zero-one Loss & 90.70\textsubscript{$\pm$0.11} & 71.79\textsubscript{$\pm$0.10} \\
    Cross Entropy & 90.53\textsubscript{$\pm$0.12} & 71.21\textsubscript{$\pm$0.13} \\
    Brier Score & 90.67\textsubscript{$\pm$0.10} & 71.79\textsubscript{$\pm$0.18} \\
    \bottomrule
    \end{tabular}
    
    \vspace{10pt}
    
    \setlength{\tabcolsep}{3pt}
    \caption{AULC of \methodname across varying numbers of retraining epochs.}\label{tab:retraining}
    \begin{tabular}{rcc}
        \toprule
        \rowcolor{gray!30}
        \textbf{\# Epochs} \hspace{25pt} & \textbf{CIFAR-10} & \textbf{DTD} \\
        \midrule
        5 & 90.00\textsubscript{$\pm$0.10} & 70.77\textsubscript{$\pm$0.11} \\
        10 & 90.57\textsubscript{$\pm$0.12} & 71.13\textsubscript{$\pm$0.13} \\
        25 & 90.71\textsubscript{$\pm$0.13} & 71.62\textsubscript{$\pm$0.12} \\
        50 & 90.70\textsubscript{$\pm$0.11} & 71.80\textsubscript{$\pm$0.11} \\
        100 & 90.67\textsubscript{$\pm$0.12} & 71.72\textsubscript{$\pm$0.11} \\
        200 (Full) & 90.60\textsubscript{$\pm$0.12} & 71.84\textsubscript{$\pm$0.07} \\
        \bottomrule
    \end{tabular}
    \end{minipage}
    
\end{table}

\subsection{Estimation of Performance}
Here, we examine how different loss functions $\ell(\cdot)$ affect the performance of \methodname. In principle, the choice of $\ell$ will influence the estimation of the performance gain of potential candidate batches. Next to the zero–one loss, which corresponds to the accuracy and therefore has direct relevance for our target metric, we also evaluate two proper scoring rules: cross-entropy and Brier score~\citep{zhao2021uncertaintyaware}. Similar to the zero–one loss, both loss functions correlate with accuracy, but also quantify the model's probabilistic calibration. As a result, they not only give insights about the performance, but also measure the reliability of the predicted probabilities.
\Cref{tab:perf_estimation} shows that zero-one loss, Brier score and cross-entropy yield similar performance, with cross-entropy slightly lagging behind. For \methodname, we opted for the zero-one loss due to its link to the accuracy, but we can equally use proper scoring rules such as the Brier score. Thus, \methodname is also suitable for scenarios were probabilistic calibration might be important. Accordingly, when employing \methodname, we recommend selecting the metric of interest appropriate for the task at hand.

\subsection{Retraining}
To lower retraining cost within \methodname, we adopt a selection-via-proxy approach, which involves assessing candidate batches by exclusively retraining the final layer for 50 epochs. The impact of different numbers of retraining epochs on the performance of \methodname is detailed in~\Cref{tab:retraining}. We see that reducing this number yields nearly identical AL performance compared to utilizing full retraining with 200 epochs. Consequently, these results suggest that a reduced number of retraining epochs during the selection is sufficient to identify effective candidate batches. Since we recognize a slight decrease in performance going from 10 to 5, and we want to ensure reliable candidate batch assessment, we select 50 retraining epochs as the default for \methodname.
However, we additionally investigate the scenario of constructing a highly efficient oracle. To this end, in \cref{app:mini-oracle}, we show how \methodname performs when both the number of candidate batches and the number of retraining epochs are greatly reduced.

\section{Limitations}
A central limitation of all oracle strategies, including \methodname, is that they should not be interpreted as upper baselines for selection strategies. In particular, the gap observed between \methodname and actual AL strategies cannot be viewed as a directly attainable performance margin. This is because oracles rely on supervised information during selection, making them infeasible in practice and fundamentally distinct from achievable benchmarks.
Consequently, the observed performance gap can be conceptually decomposed into two parts: (i) weaknesses of current AL strategies that could, in principle, be improved, and (ii) advantages that stem purely from supervised knowledge and are therefore unattainable. 
A principled decomposition of these two components remains an open research challenge. Nevertheless, the insights from our evaluation remain valuable, as a larger observed gap still plausibly indicates potential for improvement, even if the true attainable portion of this gap cannot yet be precisely quantified.

In this regard, we further emphasize that the strong performance of \methodname is not solely due to oracle knowledge, since batches are proposed by existing unsupervised AL strategies. Consequently, whenever \methodname identifies a high-performing batch, at least one underlying strategy must already have been capable of proposing it. This property suggests that, in principle, an improved selection may be conceptualized by learning to identify such high-quality batches.
To investigate the importance of supervision in \methodname, we conducted additional experiments (Appendix~\ref{app:limitations}) evaluating its performance without access to supervised strategies or ground-truth labels.

\section{Conclusion}
We introduced \methodname, an efficient oracle strategy for batch AL that scales with large datasets and complex DNNs. \methodname achieves a tractable approximation of the optimal selection by: (i) restricting the search space to candidate batches through an ensemble of selection strategies, (ii) assessing performance improvements of those batches by retraining only the final layer, and (iii) selecting batches with the highest performance improvement. 
Our experiments on ten image classification datasets demonstrate that \methodname outperforms existing oracle strategies and consistently exceeds the performance of state-of-the-art AL selection strategies. Notably, the largest performance improvements emerge on large-scale multiclass datasets, suggesting that these settings are a promising direction for research and for exploring robust, model-agnostic batch selection strategies. 
The analysis of which selection strategies were chosen showed that \methodname uses a wide range of selection strategies over several AL cycles to achieve both high performance and robustness. This suggests that future AL strategies may increasingly focus on ensemble-based approaches, which, ideally, automatically identify and apply the most appropriate selection strategy in a given cycle.

Although we focus on DNNs in this work, \methodname can easily be combined with other machine learning models. For example, kernel-based approaches are particularly well suited, as retraining can be performed easily and efficiently by updating the kernel matrix.
Furthermore, since \methodname consists of an ensemble of selection strategies, it can be easily extended to include new, state-of-the-art AL strategies.  
As a result, it will continue to provide a reliable oracle strategy in future research. In this context, we envision \methodname as a practical proxy for assessing where current strategies may still have room for improvement.
Specifically, whenever a new selection strategy is introduced, we recommend integrating it directly into the ensemble of \methodname. At the same time, we suggest to include the authors’ existing (already implemented) comparison strategies as well. This setup provides a straightforward, efficient way to establish an oracle strategy against which novel strategies can be systematically evaluated. An exemplary study demonstrating this procedure can be found in Appendix~\ref{app:practical-recomm}.

For future work, a promising direction to improve \methodname is the incorporation of a self-adaptive component that dynamically emphasizes strategies producing high-quality batches. Specifically, this could be achieved by framing the allocation as a multi-armed bandit problem, where the number of batches assigned to each strategy is adjusted based on observed batch quality. Furthermore, by measuring the similarity of selected candidate batches, we could assess redundancy among ensemble members and encourage more independent selection, increasing the likelihood of retaining diverse, informative instances. Such mechanisms would allow \methodname to automatically focus computational resources on the most effective and independent strategies for a given dataset and cycle, potentially improving both efficiency and performance.

\bibliography{main}
\bibliographystyle{tmlr}

\appendix 

\newpage
\section{Practical Recommendations}\label{app:practical-recomm}
Our oracle \methodname is best employed for evaluating and comparing AL strategies. 
It allows researchers to pose questions such as ``\textit{How far away is my newly proposed AL strategy from the optimal performance?}'' or ``\textit{At which stage of AL (e.g., early vs.~late cycles) does my strategy underperform?}''. 
If a new AL strategy is close to our oracle, it is a reliable indicator for a well working selection. Vice versa, a large gap implies that the strategy might struggle and that there may be potential for improvement.

When employing \methodname in experiments, we recommend a simple procedure without needing to implement each AL strategy listed in \cref{tab:strategies}. Specifically,  if an author has developed a novel AL strategy and intends to evaluate it alongside four additional state-of-the-art strategies, we suggest to simply use the already implemented AL strategies for candidate batch generation. Any further hyperparameters can be set to the default values presented in this work.
Regarding the budget for a given dataset, we recommend determining it by performing Random sampling until the learning curve reaches convergence.
This ensures that both low- and high-budget scenarios are accounted for in the evaluation, which can potentially reveal issues in exploration and exploitation. Furthermore, although our experiments demonstrated minimal sensitivity to the choice of loss function, employing an alternative loss function may be beneficial if the evaluation metric significantly differs from classification accuracy.

\begin{wrapfigure}{r}{.5\linewidth}
    \vspace{-10pt}
    \centering
    \includegraphics[width=\linewidth]{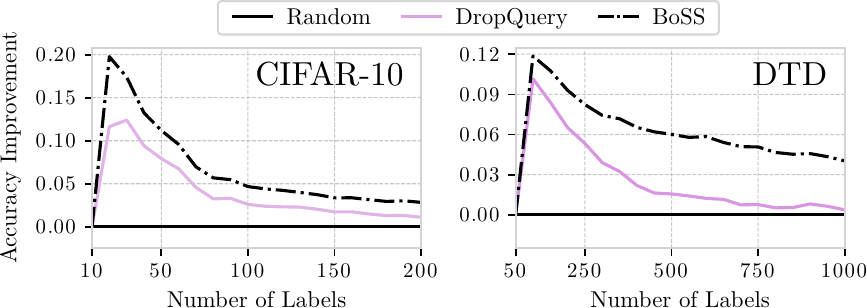}
    \caption{Explanatory plot of the ``new'' strategy DropQuery in relation to both random and \methodname.}\label{fig:example_inter}
    \vspace{-10pt}
\end{wrapfigure}

To illustrate this, we consider an experimental scenario where we assume a novel AL strategy (i.e., DropQuery) and seek to evaluate its performance against established state-of-the-art strategies (i.e., BADGE, BAIT, TypiClust).
Accordingly, we construct \methodname by defining the set $\gS$ to include DropQuery, BADGE, BAIT, TypiClust, and Random sampling, with $10$ candidate batches per strategy. 
Comparing the new AL strategy DropQuery with \methodname in \cref{fig:example_inter} provides multiple insights:
On CIFAR-10, a clear performance difference emerges in the initial cycles, after which both DropQuery and \methodname behave similarly. This suggests DropQuery might struggle to identify influential instances early on but continues to perform well for the rest of the experiment. On DTD, a more complex dataset, although the initial performance gap is smaller, the gap between DropQuery and \methodname continuously increases in subsequent cycles. This indicates that while effective in the beginning, exploitation may not working properly in later cycles.

\section{Varying Candidate Pool Size}\label{app:var_cand_pools}

\begin{wrapfigure}{r}{.55\linewidth}
    \vspace{-10pt}
    \includegraphics[width=\linewidth]{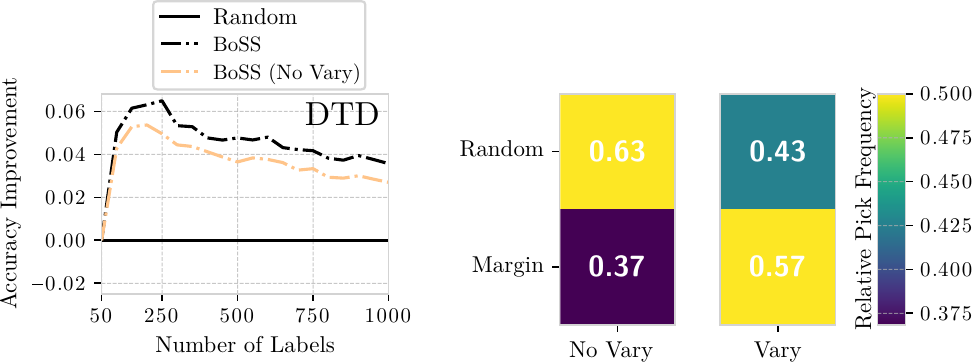}
    \caption{Relative learning curves and pick frequencies with (Vary) and without (No Vary) varying the subset size for candidate batch generation.}
    \vspace{-10pt}
    \label{fig:vary-sss}
\end{wrapfigure}
Here, we examine the influence of varying the size of the candidate pools from which candidate batches are selected. In general, sampling candidate pools enables deterministic AL selection strategies to generate multiple distinct candidate batches. Moreover, selecting batches from smaller subsets rather than from the entire unlabeled pool improves computational efficiency. However, choosing the appropriate subset size presents a trade-off. On one hand, the subset must be sufficiently large to ensure the presence of influential instances. On the other hand, overly large subsets may reduce randomness, resulting in deterministic strategies repeatedly selecting similar candidate batches.
Smaller subset sizes introduce more randomness into batch selection, whereas larger subset sizes emphasize the intrinsic characteristics of the employed AL strategies. Thus, identifying the optimal subset size is difficult.

For this reason, we vary the subset size of candidate pools in \methodname. When the subset size is small, candidate batches exhibit more randomness. In contrast, when the subset size is large, the selection of candidate batches is increasingly driven by the heuristics of the employed selection strategies. Due to the performance-based view of \methodname, we ensure that low-performing candidate batches from unsuitable subset sizes do not influence the oracles overall AL performance. 
\Cref{fig:vary-sss} demonstrates this effect. For this experiment, we include only Random and Margin as \methodname{}’s selection strategies and focus on DTD with the DINOv2-ViT-S/14 model. We observe that without varying candidate pool sizes, \methodname remains strongly biased toward randomly sampled batches. In contrast, varying pool sizes shifts selection toward Margin and yields an increase in performance. 

\section{Analysis of AL Strategies: Uncertainty vs.~Representativeness}
\begin{figure}[!ht]
    \centering
    \includegraphics[width=\linewidth]{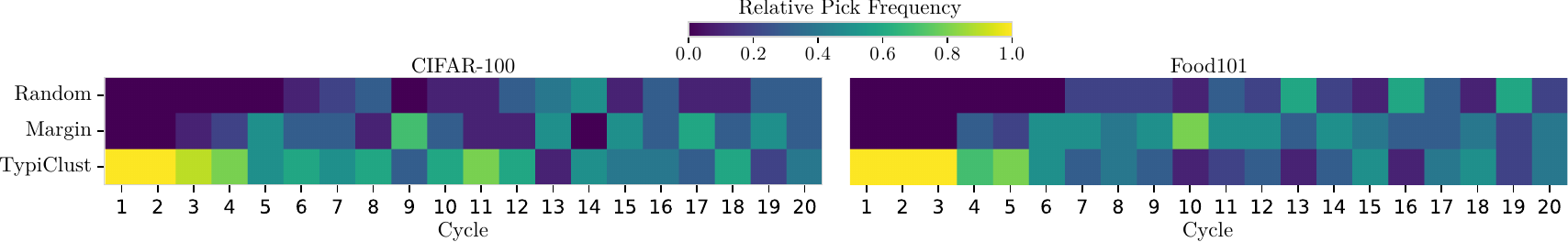}
    \caption{Average pick choices of \methodname{} with three selection strategies on CIFAR-100 and Food101.}
    \label{fig:pick-choices-repr-vs-unc-combined}
\end{figure}

In addition to its strong performance, \methodname{}’s performance-based selection of candidate batches enables us to assess which AL strategy is most effective at each cycle. Specifically, by looking at which candidate batch was selected by \methodname, we can identify whether a particular AL strategy excels early, later, or across all stages of the AL process. To illustrate this, we run \methodname{} on CIFAR-100 and Food101 with the DINOv2-ViT-S/14 model using three selection strategies for candidate batch generation. We include Random sampling, the representativeness-based strategy TypiClust, and the uncertainty-based strategy Margin. Following the intuitions from~\citep{kottke2021optimal,hacohen2023how}, early cycles should benefit from representative instances to capture the task’s underlying distribution, while later cycles should benefit from uncertain instances.
\Cref{fig:pick-choices-repr-vs-unc-combined} shows the average pick frequency of \methodname{} over ten runs on both datasets. The selection pattern reveals a clear preference for representative candidate batches in the first three cycles, as TypiClust is primarily picked at that stage. Contrary to the intuitions, however, \methodname does not exclusively focus on uncertain instances later on but continues to select a mix of random, uncertain, and representative batches. This suggests that either Margin may be less effective at identifying truly challenging instances on these datasets or that the intuition that AL should pivot solely to uncertain instances may be overly simplistic. Furthermore, the fact that randomly sampled candidate batches are chosen suggests that none of the selection strategies provide influential batches at a given stage. 

\section{Pick Choices Per Datasets}\label{app:picks_per_dataset}
\begin{figure}[!ht]
    \centering
    \includegraphics[width=\linewidth]{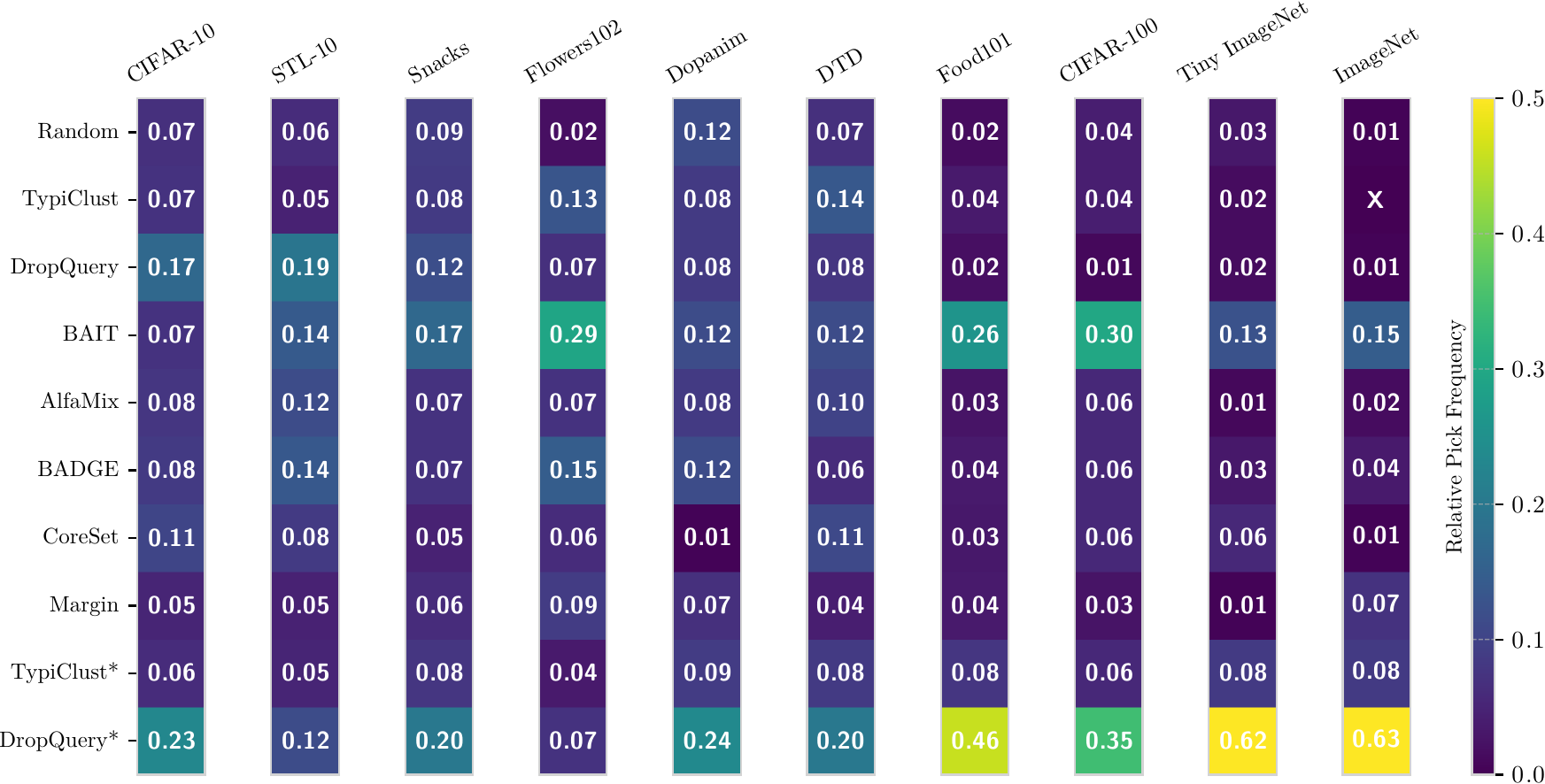}
    \caption{Relative pick frequencies of selection strategies by \methodname per dataset averaged over cycles, here given with the numeric value (in \%).}
    \label{fig:batch-sel-per-dset}
\end{figure}
Supplementing \hyperlink{rq4}{RQ\textsubscript{4}}, we present the average relative pick frequencies for each dataset. \Cref{fig:batch-sel-per-dset} shows how often each selection strategy was chosen as the best candidate batch across datasets. Three main insights emerge.
First, although DropQuery* and BAIT achieve the highest pick frequencies on several datasets, every selection strategy contributes influential candidate batches at various stages of the AL process. This underscores the value of including each selection strategy in \methodname.
Second, not only does every selection strategy get selected at least once, confirming that no single strategy dominates an entire AL cycle, but there is also no consistently preferred selection strategy across all datasets. This highlights that the ensemble of AL strategies itself is critical for maintaining strong, dataset-agnostic performance. Third, for datasets with larger batch sizes ($\ge 100$), pick frequencies mostly concentrate on two selection strategies. This pattern suggests that the other AL strategies struggle to propose effective candidate batches as dataset complexity grows. Moreover, since DropQuery* cannot be simply applied in AL (labels in the unlabeled pool are unavailable), BAIT emerges as a promising alternative in that context.

\section{Minimal Oracle}\label{app:mini-oracle}
\begin{figure}[!ht]
    \centering
    \includegraphics[width=.9\linewidth]{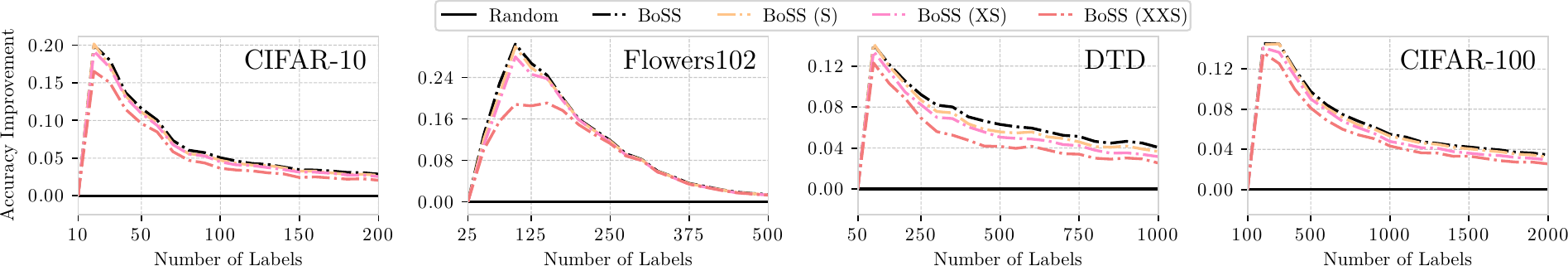}
    \caption{Relative learning curves of \methodname and its runtime-optimized variants.}
    \label{fig:batch-sel-per-dset-mini}
\end{figure}

In \Cref{sec:ablations}, we investigated various factors influencing the performance of \methodname and settled for a good trade-off between runtime and effectiveness. Here, we aim to examine how \methodname{}'s performance decreases when prioritizing runtime only. Therefore, we introduce three different runtime-optimized variants of our original oracle, namely \methodname (S) with $T=50$ and $25$ retraining epochs, \methodname (XS) with $T=25$ and $10$ retraining epochs, and \methodname (XXS) with $T=10$ and $5$ retraining epochs. The results in \cref{fig:batch-sel-per-dset-mini} show that while these runtime-optimized variants yield slightly reduced performance, \methodname still performs reasonably. Especially considering the simpler dataset CIFAR-10, even \methodname (XXS) still yields the best performance when compared to all considered state-of-the-art AL strategies. 
Thus, we want to emphasize that the values chosen in \Cref{sec:ablations} are guideline values and that \methodname can also work well when runtime needs to be significantly reduced.

\section{Supplementary Absolute Learning Curves}\label{app:lcs}
\begin{figure}[!htp]
    \centering
    \includegraphics[width=\linewidth]{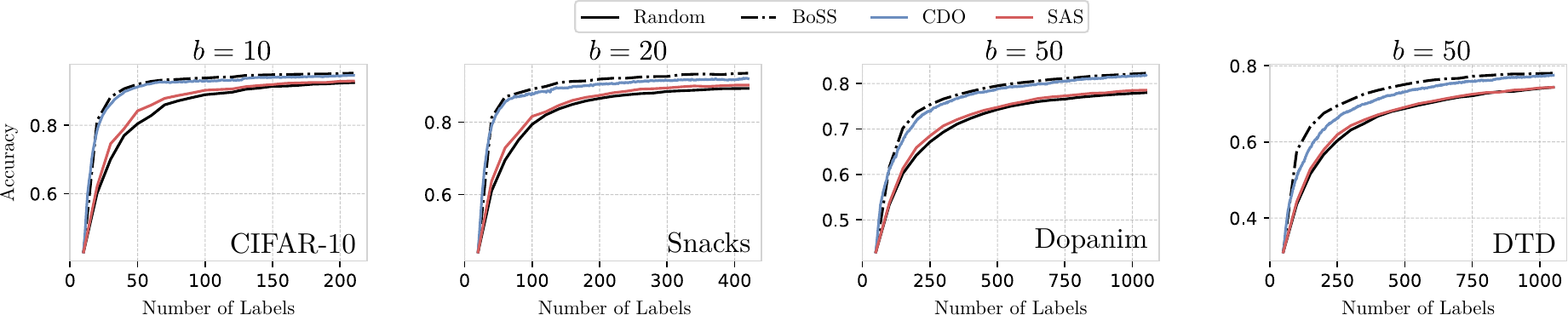}
    \caption{Absolute learning curves of oracle strategies with aligned runtimes using DINOv2-ViT-S/14.}
    \label{fig:absolute_oracles}
\end{figure}

\begin{figure}[!htp]
    \centering
    \includegraphics[width=\linewidth]{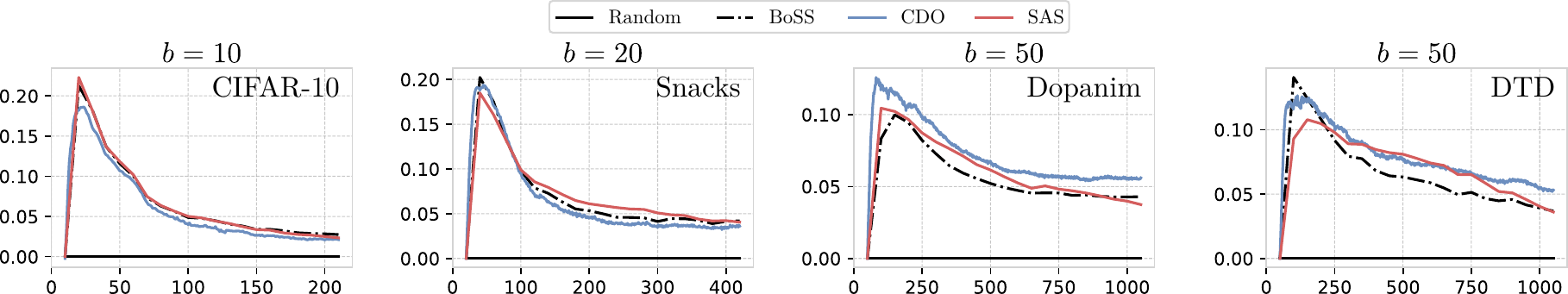}
    \caption{Relative learning curves of oracle strategies with default hyperparameters using DINOv2-ViT-S/14.}
    \label{fig:default_oracles}
\end{figure}

\begin{figure}[!htp]
    \centering
    \subfloat[DINOv2-ViT-S/14]{\includegraphics[width=\linewidth]{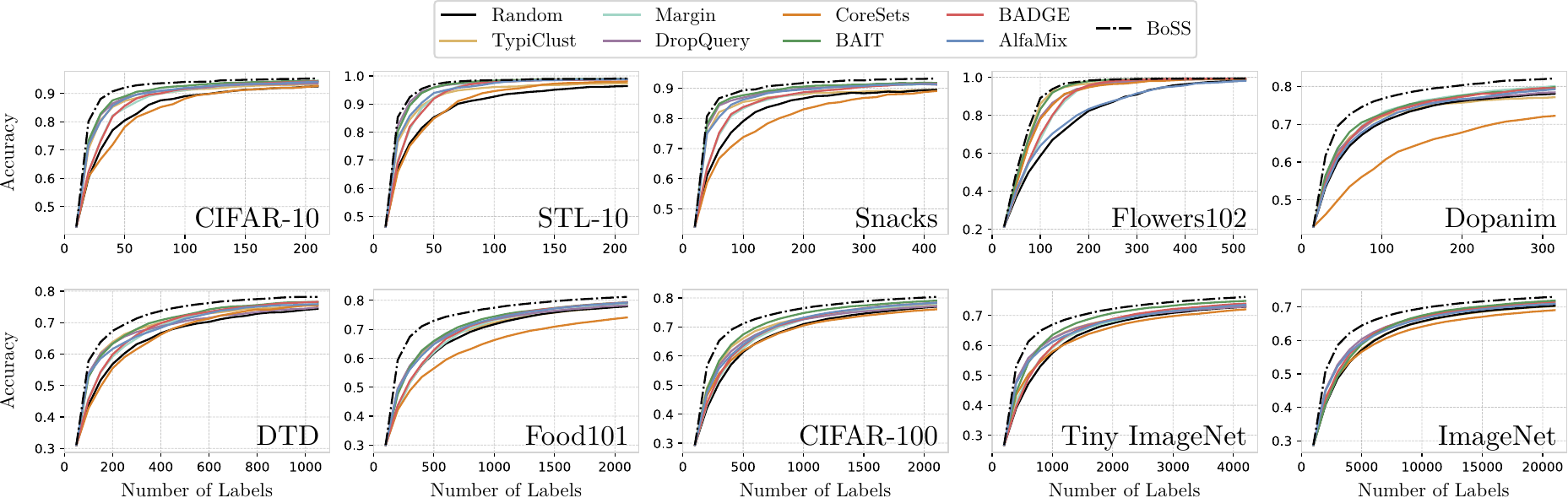}}
    \vspace{1em}
    \subfloat[SwinV2-B]{\includegraphics[width=\linewidth]{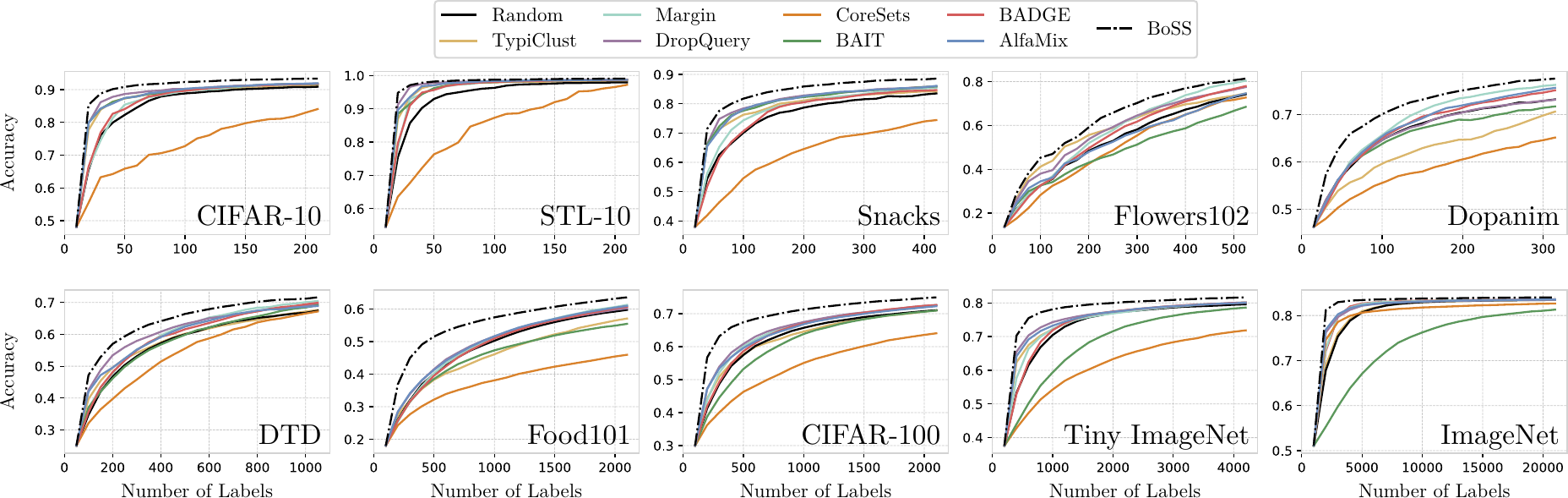}}
    \caption{Absolute learning curves achieved by \methodname and state-of-the-art selection strategies at each annotation cycle for different pretrained models.
    }
    \label{fig:absolute_sota}
\end{figure}

In addition to the relative learning curves presented in the main part of the paper, we also show the associated absolute learning curves here. \Cref{fig:absolute_oracles} depicts the absolute learning curves that correspond to the relative curves reported in \cref{fig:oracle_experiments}. Similarly, in \cref{fig:absolute_sota} we report the corresponding absolute learning curves of the state-of-the-art experiments from \cref{fig:sota_experiments}. Additionally, we also report the absolute learning curves of all oracle strategies with default hyperparameters in~\cref{fig:default_oracles}.

\section{Budget-Regime Analysis with DINOv2}\label{app:budget_analysis}
Here, we additionally analyze the area under learning curves (AULC) across all budget regimes for the DINOv2-ViT-S/14 backbone with to goal to facilitate an easy comparison to related work. Note that the reported AULC values correspond to specific segments of the learning curve: low (cycles 1--7), mid (cycles 7--14), and high (cycles 14--20), allowing for a more granular assessment of performance at different annotation budgets.

We observe clear differences between the selection strategies across all budgets. At \textbf{low-budget} (\cref{tab:aulc_low}), performance differences are substantial. Simple baselines like Random or CoreSets yield weak results, while more sophisticated strategies such as DropQuery, BAIT, and \methodname clearly outperform them across datasets. 
At \textbf{mid budget} (\cref{tab:aulc_mid}), the gap between strategies narrows, but \methodname remains the leading approach, followed by BAIT and BADGE. Margin and TypiClust provide moderate improvements over Random sampling.
At \textbf{high budget} (\cref{tab:aulc_high}), performances saturate and differences between strategies become smaller. Nevertheless, \methodname consistently yields the highest AULC values across all datasets, with BAIT and BADGE being strong feasible strategies. 
\textbf{Overall}, \methodname dominates in all budget regimes, highlighting its role as an oracle. Among practically usable strategies, BAIT and BADGE are the most competitive, especially at medium and high budgets, while CoreSets tends to underperform throughout.
\begin{table}[!ht]
    \centering
    \caption{AULC for the low-budget regime across vision datasets using DINOv2 features.}\label{tab:aulc_low}
    \scriptsize
    \setlength{\tabcolsep}{4pt}
    \begin{tabular}{lrrrrrrrrrr}
    \toprule
     & CIFAR-10 & STL-10 & Snacks & Flowers102 & Dopanim & DTD & Food101 & CIFAR-100 & Tiny ImageNet & ImageNet \\
    \midrule
    Random & 0.713 & 0.763 & 0.707 & 0.548 & 0.612 & 0.531 & 0.538 & 0.531 & 0.496 & 0.500 \\
    Margin & 0.745 & 0.813 & 0.740 & 0.630 & 0.628 & 0.550 & 0.539 & 0.544 & 0.513 & 0.513 \\
    CoreSets & 0.693 & 0.759 & 0.668 & 0.680 & 0.525 & 0.519 & 0.503 & 0.543 & 0.508 & 0.501 \\
    BADGE & 0.748 & 0.810 & 0.742 & 0.640 & 0.626 & 0.555 & 0.546 & 0.550 & 0.512 & 0.519 \\
    TypiClust & 0.777 & 0.825 & 0.776 & 0.721 & 0.629 & 0.587 & 0.571 & 0.577 & 0.546 & 0.503 \\
    DropQuery & 0.789 & 0.862 & 0.792 & 0.693 & 0.629 & 0.584 & 0.576 & 0.567 & 0.546 & 0.530 \\
    BAIT & 0.796 & 0.857 & 0.799 & 0.717 & 0.638 & 0.586 & 0.575 & 0.584 & 0.545 & 0.511 \\
    AlfaMix & 0.783 & 0.836 & 0.780 & 0.580 & 0.617 & 0.573 & 0.569 & 0.561 & 0.537 & 0.523 \\
    \methodname & 0.829 & 0.875 & 0.813 & 0.744 & 0.677 & 0.619 & 0.643 & 0.626 & 0.587 & 0.569 \\
    \bottomrule
    \end{tabular}
\end{table}
\begin{table}[!ht]
    \centering
    \caption{AULC for the mid-budget regime across vision datasets using DINOv2 features.}\label{tab:aulc_mid}
    \scriptsize
    \setlength{\tabcolsep}{4pt}
    \begin{tabular}{lrrrrrrrrrr}
    \toprule
     & CIFAR-10 & STL-10 & Snacks & Flowers102 & Dopanim & DTD & Food101 & CIFAR-100 & Tiny ImageNet & ImageNet \\
    \midrule
    Random & 0.893 & 0.931 & 0.869 & 0.889 & 0.747 & 0.695 & 0.725 & 0.715 & 0.681 & 0.662 \\
    Margin & 0.916 & 0.979 & 0.888 & 0.980 & 0.765 & 0.722 & 0.736 & 0.732 & 0.687 & 0.674 \\
    CoreSets & 0.886 & 0.949 & 0.835 & 0.967 & 0.658 & 0.701 & 0.671 & 0.711 & 0.669 & 0.647 \\
    BADGE & 0.919 & 0.979 & 0.888 & 0.981 & 0.760 & 0.726 & 0.744 & 0.736 & 0.695 & 0.677 \\
    TypiClust & 0.913 & 0.961 & 0.882 & 0.981 & 0.748 & 0.721 & 0.729 & 0.736 & 0.693 & 0.662 \\
    DropQuery & 0.919 & 0.981 & 0.900 & 0.973 & 0.756 & 0.709 & 0.740 & 0.731 & 0.692 & 0.675 \\
    BAIT & 0.930 & 0.984 & 0.906 & 0.988 & 0.763 & 0.732 & 0.750 & 0.753 & 0.712 & 0.680 \\
    AlfaMix & 0.922 & 0.976 & 0.898 & 0.893 & 0.753 & 0.718 & 0.744 & 0.738 & 0.690 & 0.670 \\
    \methodname & 0.942 & 0.986 & 0.919 & 0.989 & 0.793 & 0.756 & 0.779 & 0.769 & 0.727 & 0.700 \\
    \bottomrule
    \end{tabular}
\end{table}

\begin{table}[!ht]
    \centering
    \caption{AULC for the high-budget regime across vision datasets using DINOv2 features.}\label{tab:aulc_high}
    \scriptsize
    \setlength{\tabcolsep}{4pt}
    \begin{tabular}{lrrrrrrrrrr}
    \toprule
     & CIFAR-10 & STL-10 & Snacks & Flowers102 & Dopanim & DTD & Food101 & CIFAR-100 & Tiny ImageNet & ImageNet \\
    \midrule
    Random & 0.919 & 0.959 & 0.889 & 0.972 & 0.774 & 0.734 & 0.770 & 0.759 & 0.720 & 0.696 \\
    Margin & 0.936 & 0.988 & 0.910 & 0.993 & 0.793 & 0.760 & 0.777 & 0.771 & 0.726 & 0.707 \\
    CoreSets & 0.919 & 0.977 & 0.879 & 0.989 & 0.710 & 0.746 & 0.726 & 0.751 & 0.710 & 0.682 \\
    BADGE & 0.938 & 0.987 & 0.911 & 0.993 & 0.789 & 0.759 & 0.784 & 0.775 & 0.731 & 0.708 \\
    TypiClust & 0.930 & 0.971 & 0.892 & 0.990 & 0.768 & 0.749 & 0.770 & 0.766 & 0.725 & 0.695 \\
    DropQuery & 0.933 & 0.986 & 0.911 & 0.990 & 0.778 & 0.740 & 0.774 & 0.763 & 0.722 & 0.704 \\
    BAIT & 0.942 & 0.989 & 0.916 & 0.993 & 0.790 & 0.761 & 0.784 & 0.783 & 0.742 & 0.711 \\
    AlfaMix & 0.937 & 0.987 & 0.912 & 0.969 & 0.783 & 0.754 & 0.783 & 0.775 & 0.726 & 0.702 \\
    \methodname & 0.951 & 0.990 & 0.930 & 0.993 & 0.816 & 0.778 & 0.805 & 0.797 & 0.754 & 0.724 \\
    \bottomrule
    \end{tabular}
\end{table}

\section{Additional Results on Text Classification Tasks}\label{app:text_results}
\begin{figure}[!ht]
    \centering
    
    \subfloat[Absolute learning curves.]{\includegraphics[width=\linewidth]{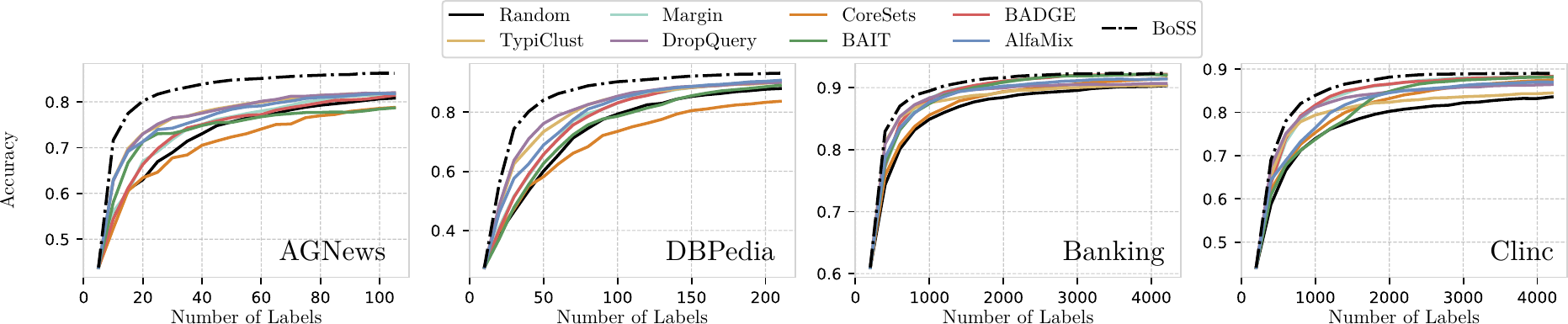}}
    \vspace{1em}
    \subfloat[Relative learning curves.]{\includegraphics[width=\linewidth]{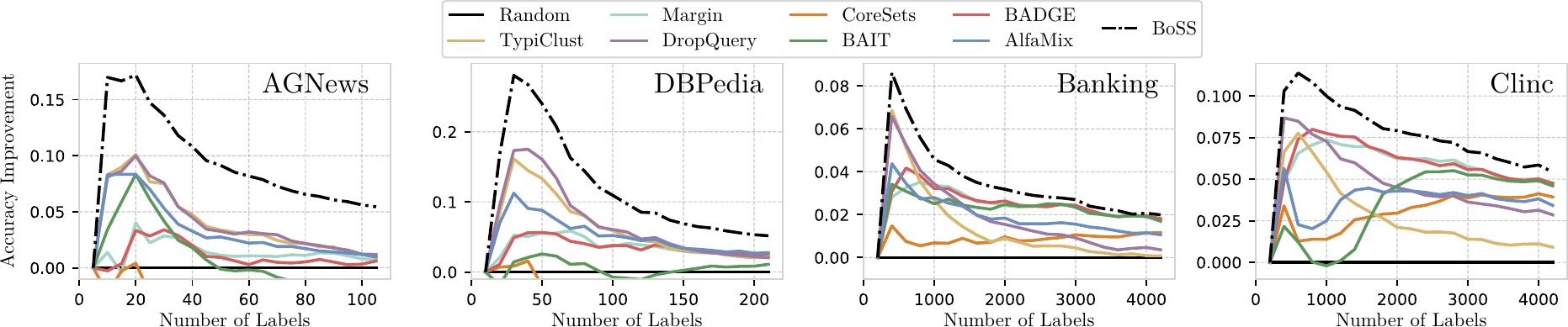}}
    
    \caption{Learning curves of \methodname and state-of-the-art selection strategies across annotation cycles on four text classification datasets using MiniLM features.}
    \label{fig:text_results}
\end{figure}

We additionally evaluated \methodname on four text classification datasets oriented on the text AL benchmark~\citep{rauch2023activeglae}. 
In particular, we include the large-scale news classification dataset AGNews~\citep{zhang2015agnews} ($K=4$), the medium-cardinality ontology classification dataset DBPedia~\citep{lehmann2015dbpedia} ($K=14$), and the high-cardinality intent detection dataset Banking~\citep{casanueva2020banking} ($K=77$). 
Furthermore, to extend the evaluation beyond 100 classes, we also include the intent classification dataset Clinc~\citep{larson2019clinc} ($K=150$). 
Similar to the setup in the main paper, we extracted features using a pretrained model. Here, we employ the MiniLM language model~\citep{reimers2019sentencebert}, though any model pretrained on text data could have been used to extract features. 
\Cref{fig:text_results} reports absolute and relative learning curves over 10 repetitions. 
The results confirm that \methodname also achieves strong performance in the text domain, reinforcing its applicability beyond vision.

\section{Analyzing the Impact of Supervision in \methodname}\label{app:limitations}
To analyze the effect of supervision within \methodname, we conduct two controlled variants that progressively reduce the amount of supervised information available during batch selection. 

\textbf{\methodname without supervised strategies:} In this setting, we restrict the candidate batches to only be generated by unsupervised AL strategies, i.e., methods that do not rely on label information at selection time. This allows us to examine the performance of \methodname persists when the supervised signal is entirely removed from the candidate generation process. 
\textbf{\methodname with approximate supervision:} Here, instead of using ground-truth labels for retraining or performance evaluation, we infer them by an approximate distribution. Specifically, we transform the original formulation in \cref{eq:batch_perf} to be usable with no labels
\begin{align}
    \argmin_{\gB\subset\gU}  \E_{p(\vx, y)} \big[\ell\big( y, p(y |\vx, \gL^+)\big)\big] 
    &\approx \argmin_{\gB\subset\gU} \E_{p(\vy_\gB |\mX_\gB)} \E_{p(\vx)} \E_{p(y|\vx)}\big[\ell\big( y, p(y |\vx, \gL^+)\big)\big] \\
    &\approx  \argmin_{\gB\subset\gU} \E_{p(\vy_\gB |\mX_\gB, \vtheta^\star)}  \E_{p(\vx)} \E_{p(y|\vx, \vtheta^\star)}\big[\ell\big( y, p(y |\vx, \gL^+)\big)\big],
\end{align}
where we approximate $p(\vy_\gB |\mX_\gB)$ and $p(y|\vx)$ with a model $\theta^\star$ trained on the entire training dataset. This approximation represents a more realistic oracle that relies on predictive uncertainty rather than perfect supervision. We employ the DINOv2-ViT-S/14 backbone with the same experimental setup as in the main paper.

\begin{figure}[!ht]
    \centering
    \subfloat[Absolute learning curves.]{\includegraphics[width=\linewidth]{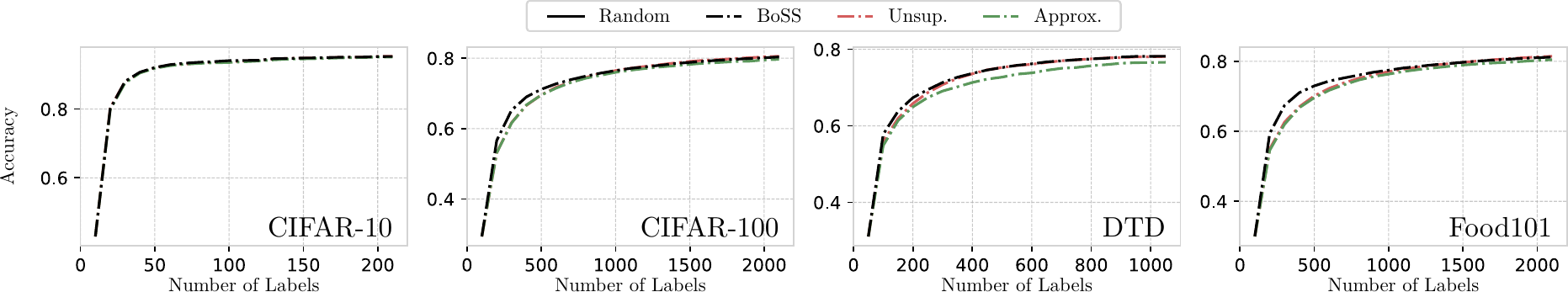}}
    \vspace{1em}
    \subfloat[Relative learning curves.]{\includegraphics[width=\linewidth]{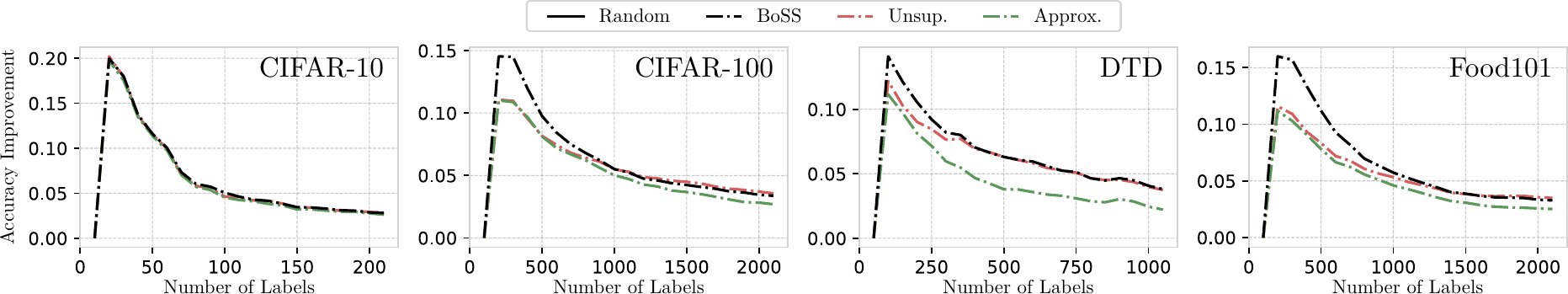}}
    \caption{Learning curves of \methodname and its variants with reduced supervision: without supervised strategies (Unsup.) and with approximate supervision (Approx.).}
    \label{fig:limitations}
\end{figure}
Based on the learning curves in \cref{fig:limitations}, we see that the performance gap between different supervision levels of \methodname varies across datasets. On CIFAR-10, both \methodname without supervised strategies (Unsup.) and the approximate variant (Approx.) achieve accuracies comparable to the fully supervised oracle, indicating that label information provides minimal additional benefit. However, on more challenging datasets with many classes (CIFAR-100 and Food101), supervised strategies seem to play a more important role, particularly during the early stages of AL. However, by the end of the AL process, the unsupervised variant is on par with the fully supervised oracle, suggesting that the value of supervision diminishes as the labeled pool grows. 
Notably, on DTD, a dataset where DINOv2-ViT-S/14 appears less well-suited due to substantial domain mismatch, the approximate supervision variant (Approx.) exhibits degraded performance. We believe this is due to areas with high aleatoric uncertainty in the embedding space that yield incorrect labels. To potentially mitigate this, we suggest employing predictive models that explicitly disentangle uncertainties.
Overall, our experiment shows that supervision can play an important role in \methodname when datasets are noisy or exhibit significant domain shift. For future work, we believe that learning a candidate batch selection policy represents a promising research direction. This way, it might be possible to imitate \methodname's behavior by directly predicting the most promising candidate batch without requiring explicit label access.

\end{document}